\pdfoutput=1
\documentclass{article}

\usepackage{bm}
\usepackage[boldmath]{numprint}
\usepackage[inkscapeformat=png]{svg}
\usepackage{amssymb}
\usepackage{authblk}
\usepackage{comment}
\usepackage{arxiv}
\usepackage[utf8]{inputenc} % allow utf-8 input
\usepackage[T1]{fontenc}  % use 8-bit T1 fonts
\usepackage{hyperref} % hyperlinks
\usepackage{url} % simple URL typesetting
\usepackage{booktabs} % professional-quality tables
\usepackage{amsfonts} % blackboard math symbols
\usepackage{nicefrac} % compact symbols for 1/2, etc.
\usepackage{microtype} % microtypography
\usepackage{lipsum}
\usepackage{graphicx}
\graphicspath{ {./pic/} }
\usepackage{amsmath} 
\usepackage{multirow}

\newcount\Comments % 0 suppresses notes to selves in text
\usepackage{color}
\definecolor{darkgreen}{rgb}{0,0.5,0}
\definecolor{purple}{rgb}{1,0,1}
\definecolor{orange}{rgb}{1.0, 0.6, 0.4}
\newcommand{\kibitz}[2]{\ifnum\Comments=1\textcolor{#1}{#2}\fi}

\newcommand{\kizito}[1]{\kibitz{red} {[Kizito: #1]}}

\title{\kizito{Data-driven? }Derivative-based regularization for Neural Networks}
\title{Derivative-based Regularization for Regression}
  
\author[1]{Enrico Lopedoto \texttt{enrico.lopedoto@city.ac.uk}} 
\author[2]{\\Maksim Shekhunov  \texttt{shexunov@gmail.com}} 
\author[1]{\\Vitaly Aksenov \texttt{v.aksenov@city.ac.uk}}
\author[1]{\\Kizito Salako \texttt{k.o.salako@city.ac.uk}}
\author[1]{\\Tillman Weyde \texttt{t.e.weyde@city.ac.uk}}

\affil[1]{Department of Computer Science\\
 City University London\\
 London, UK\\}
\affil[2]{Department of Computer Science\\
 ITMO University\\
 St Petersburg, Russian Federation}

\begin{document}

\maketitle

\begin{abstract}
In this work, we introduce a novel approach to regularization in multivariable regression problems. 
Our regularizer, called \emph{DLoss}, penalises differences between the model's derivatives and derivatives of the data generating function as estimated from the training data. %\kizito{no, not derivatives \emph{of data points}.}\tillman{fixed}\kizito{seen and commented out} 
We call these estimated derivatives \emph{data derivatives}. 
The goal of our method is to align the model to the data, not only in terms of target values but also in terms of the derivatives involved. 
To estimate data derivatives, we select (from the training data) 2-tuples of input-value pairs,  %\kizito{``tuples of pairs'' is unclear at this point; the reader does not know what these are.}\tillman{fixed}\kizito{seen and commented out} 
%Because is not efficient to choose all the possible tuples, for each provided data point we take only 1 or 3 tuple based on some 
using either nearest neighbour or random, selection. 
%\kizito{This (implementation) detail seems unnecessary for the abstract. Focus on main contributions of the work.}\tillman{fixed}\kizito{seen and commented out} 
On synthetic and real datasets, we evaluate the effectiveness of adding \emph{DLoss}, with different weights, %\kizito{unclear what coefficients you are referring to}\tillman{fixed}\kizito{seen and commented out}
to the standard mean squared error loss. 
The experimental results show that with \emph{DLoss} (using nearest neighbour selection) we obtain, on average, the best rank with respect to MSE on validation data sets, compared to no regularization, L2 regularization, and Dropout.%\kizito{Improve abstract's grammar, punctuation, etc. Should I do a pass?}\tillman{yes, please}\kizito{seen and commented out} 
\end{abstract}

\keywords{Machine learning, neural networks, regularization, regression, DLoss, data derivative
%\kizito{At some point, choose more relevant keywords. For example, we don't deal with differential equations.}\enrico{True, perhaps any suggestions that could help me understand?}\tillman{keywords updated}\kizito{seen and commented out}
}

\section{Introduction}
\label{introduction}

Regularization is used by many different methods in statistics, inverse problems, and machine learning, to prevent overfitting and numeric instability when a model is fitted to data \cite{Kukacka2017regsurvey}.
Traditional regularization methods use an additional term in the loss function used during model training, where this additional term is based on the model parameters. 
Most frequently used are the $L_1$ \cite{tibshirani1996L1} and $L_2$~\cite{TikhonovL2} norms of model parameters (i.e., norms of the coefficients of a linear model or the weights of a neural network). 
Another widely used regularization method (but only for neural networks) is \emph{DropOut}, where there is no additional loss term while neurons are randomly switched off during training~\cite{hinton2012L2,srivastava2014dropout}.

The distinguishing feature of all these regularizers is that they do not explicitly use characteristics of the unknown target function, where these characteristics may be inferred from training data. %\kizito{I know what we mean by ``independent'', but it may give the wrong impression that these regularizers do not use data, which they do. How about ``...is that they do not explicitly use characteristics of the unknown target function, where these characteristics may be inferred from training data''?}. \enrico{nice, adopted your suggestion}\kizito{seen nd commented out}
In this paper, we propose a novel regularization approach for regression, called \emph{DLoss}, that (a) is derived from training data, and (b) is applicable to any type of model where one can specify a loss function.
\emph{DLoss} is formulated as a term in the loss function that penalizes the difference between a model derivative and a derivative estimated from data points. 

Our contributions are:
\begin{enumerate}
 \item We propose a new regularization method for regression models, \emph{DLoss};
 \item We propose two variants of \emph{DLoss} using nearest neighbour or random selection of data points;
 \item We provide a publicly available implementation in PyTorch;
 \item We evaluate our approach on different datasets and compare it to training with MSE without regularization, with $L_2$ regularization, and with Dropout.
\end{enumerate}
The results show that \emph{DLoss} leads to consistently better generalisation results at a moderate increase of the computational cost.

The paper is structured as follows.
In \autoref{sec:related}, we discuss related work on regularization and differential equations. 
In \autoref{sec:method}, we introduce our method. 
In \autoref{sec:experiments}, we describe the setup of our experiments. 
In \autoref{sec:results}, we present and discuss the experimental results, and in \autoref{sec:conclusion} we draw conclusions and discuss possible future work.

%%%%%%%%%%%%%%%%%%%%%%%%%%%%%%%%%%%%%

\section{Related Work}\label{sec:related}

Our method uses information derived form the data and is not specific to any model type as opposed to common regularization approaches.
Since we are using the derivatives in our approach we also overview the related work on Neural Differential Equations.

\subsection{Regularization}
Regularization aims to reduce overfitting \cite{Kukacka2017regsurvey}.
Common regularization methods, $L_1$ and $L_2$ loss on the parameters, are also called \emph{weight decay} when applied to neural networks.
In generalised linear models and neural networks, lower weights lead to smaller gradients of the model function with respect to the inputs. 

Intuitively speaking, smaller gradients mean that flat and smooth functions are preferred. 
This preference is independent of the training data.

\emph{DropOut} is another popular regularization technique and is specific to neural networks \cite{baldi2013understanding}.
It works by switching off neurons at random during training~\cite{srivastava2014dropout}. 

\subsection{Neural Networks, Derivatives and Differential Equations} 

There are various types of models that integrate differential equations with Neural networks. 

%\kizito{This section contains a number of incorrect assertions about NDEs, and gives the impression that we are naive and don't really know what they are. Moreover, statements are associated with citations that do not support the statements. Another editorial pass is needed here, that is technically accurate, and presents ideas using clear formal language. For example, what is the technical rationale for NDEs, what differential information do they use/approximate, what is the outcome of using such information, and how do these (i.e. rationale, differential information, training outcome) differ from our approach?}\tillman{@Kizito, could you please point out any incorrect statements, help fix them and provide a perspective on the relationship between NDEs and our approach?}
%\kizito{The following prose is intended to detail: 1) the technical rationale for NDEs; 2) what differential information is used; 3) what is the output of an NDE after training; 4) the source of derivatives and what we do with them; and 5) how are these different from our approach}
%\tillman{I've done a short version, added some structure. I think this should be OK now, if not let's remove any problematic bits.}\kizito{For those papers I have read, this prose works.}\enrico{thus commented}

%\paragraph{Approximating derivatives of known functions}\kizito{The spacing with the paragraph command looks weird} 
\textbf{Approximating derivatives of known functions} by neural networks 
was already studied by \cite{hornik-et-al-1990-universal} who showed that neural networks can approximate functions and their derivatives. 
This approach assumes that the derivatives of the approximated function are known, and a modification of the network architecture is required in order to achieve the approximation. 
A more recent approach to approximating a known function with neural networks has been presented in \cite{Avrutskiy2017}, which operates similar to our approach and does not require a change to network architectures. 
It extended to higher-order derivatives in 
\cite{Avrutskiy2021}. 

%\paragraph{Integrating prior knowledge} \kizito{The spacing with the paragraph command looks weird}
\textbf{Integrating prior knowledge} by using derivatives was, to the best of our knowledge, first introduced in
\cite{LeCun91} as \emph{tangent prop}. 
Tangent prop integrates prior knowledge about invariances into the network: the network is trained using desired directional derivatives of a target function with respect to the changes in the inputs. 
This is used in the context of image classification to encourage small derivatives in directions where changes of the inputs should not affect the output. 
For example, for handwritten digits, rotations by a small amount should not affect the class output. 
Tangent prop results in a fitted neural network that approximates the desired differential behaviour. 

This approach has, in recent years, been re-kindled in the more general framework of physics-informed neural networks (PINNs) \cite{raissi2019physics,cuomo2022scientific}, which injects knowledge from physics into neural networks, in the form of differential equations. 

%\paragraph{Numeric solutions to differential equations}
\textbf{Numeric solutions to differential equations} by neural networks have been used since the 1990s. 
Early work by \cite{lee1990neural} already introduced this approach, more recent examples are \cite{parisi2003solving} and \cite{malek2006numerical}, and an overview can be found in \cite{beck2023overview}.

%\paragraph{Neural differential equations}
\textbf{Neural differential equations} (NDE) are a more recent approach to unify neural networks with differential equations (for an overview, see \cite{Kidger22}).
A neural ordinary differential equation (NODE) \cite{NEURIPS2018_Chen} estimates a continuous-time function that defines an ordinary differential equation (ODE), where this ODE approximates the discrete-time behaviour of a recursive process (such as an RNN). In essence, the recursive mapping of inputs to outputs is approximated by motion along the solution to the ODE~--- i.e., motion along a vector field.
Neural stochastic differential equations (NSDE) \cite{tzen2019neural} define diffusion processes, rather than vector fields. 
They are also used to learn continuous-time models, but of \textit{stochastic} recurrent processes. 
%So, for NODEs, one assumes that an ODE characterises the change between inputs and outputs at a given time step for, say, an RNN. 
In contrast, for our work, we assume the target function is continuously differentiable and its differential characteristics can be estimated from data. 

%During the training process, our model is encouraged to match these characteristics estimated from training data. Where NODEs use derivatives of a (recursive) network's hidden-layer outputs with respect to time, \emph{DLoss} uses derivatives of a feed-forward network's outputs with respect to the network's inputs. 

%\kizito{This paragraph mixes two related applications. \cite{Avrutskiy2021} improves the accuracy of an FFNN by encouraging the network, during training, to match the outputs of an unknown target function, as well as to learn known differential characteristics of the target. While \cite{Avrutskiy2017} generalises back-propagation to include higher order derivatives. Applications of the ideas presented in these papers can be used when approximating the solution of differential equations using neural networks. Note that these do not fall under NDEs, strictly speaking.} \enrico{changed to text, please review and comment}

%%%%%%%%%%%%%%%%%%%%%%%%%%%%%%%%%%%%%

\section{Derivative-based Regularization Method}\label{sec:method}

In this study, we introduce a regularization term that aims to align model derivatives with estimated derivatives of the target function.
This approach is data driven, without imposing a prior on the model parameter weights. %\kizito{which weights? Model parameter weights? Loss function weights?}\enrico{model parameter - updated for clarity, thanks}\kizito{seen and commented out}
%\kizito{shortedned the following to reduce redundacny}\enrico{nice - thanks}\kizito{seen and commented out}
%Notwithstanding the increasing use of NDE, the use of the training data to estimate the derivative of the target function and align it to the derivative of the model during training is, to the best of our knowledge, proposed here for the first time.
Notwithstanding the increasing use of NDEs, this form of regularization is, to the best of our knowledge, proposed here for the first time.

\subsection{Intuition}
Our regularizer aligns model derivatives with estimated derivatives of the unknown, true, target function that generated the data.
By considering derivatives of the target (estimated from training data), in addition to the target values, our approach seems intuitively more promising than other regularization approaches that rely solely on target values, and more generally applicable than approaches that rely on prior knowledge of the target's differential characteristics.

For our regularizer, the following information is needed.
Firstly, it needs derivatives of the model with respect to the input.
These derivatives may be calculated analytically, or can be approximated by a finite difference approach given the model.
Secondly, it needs estimates of target function derivatives with respect to the input using a simple differentiation formula. 
We want to minimise the difference between the model derivative and the estimated target derivative; i.e., we want to minimise the value of a \emph{DLoss} function, which we will define.

\subsection{Regression}
\label{subsec_regressiontasks}

We focus on \emph{regression problems}~--- that is, the approximation of an unknown, continuous, real-valued target function with $k$ arguments, $g(\cdot):{\mathbb R}^{k}\to{\mathbb R}$, by a model, $f(\cdot,\boldsymbol{\beta}):{\mathbb R}^{k}\to{\mathbb R}$, parameterised by a vector $\boldsymbol{\beta}\in{\mathbb R}^m$. 
Using training data consisting of input-output pair $(\mathbf{x}_i,y_i)$ (for $i=1,\ldots,n$, where $n$ is the number of training data points), regression modeling aims to optimise the parameter vector $\boldsymbol{\beta}$ of this function $f$, so that for general input-output pair $(\mathbf{x},y)$, 
\begin{equation}
 f(\mathbf{x},\boldsymbol{\beta}) = \hat{y}
\end{equation}
approximates the unknown function $g(\mathbf{x})=y$ sufficiently well. 
The variable $\mathbf{x}\in\mathbb{R}^k$ is called an 
%\textit{regressor} or 
%TW removed regression, as we use it later for regression model.
%KS okay, but this non-standard use of the term regressor could be confusing
\textit{independent variable vector} in statistics, and called \textit{features} or \textit{input vectors} in machine learning. The corresponding $y \in \mathbb{R}$, that satisfies $g(\mathbf{x})=y$, is the \textit{dependent variable}, \textit{response variable} or \textit{target}.
In this paper, we refer to an input vector $\mathbf{x}$ as a point, $(\mathbf{x},y)$ as a pair, and two pairs $((\mathbf{x}_i,y_i),(\mathbf{x}_j,y_j))$ as a tuple.

For a given parameter vector $\boldsymbol{\beta}$, the model makes a prediction, $\hat{y}_i = f(\mathbf{x}_i,\boldsymbol{\beta})$, for the $i$th training input $\mathbf{x}_i$.
In general, it is expected that $\hat{y}_i$ will differ from the target value $y_i$. The task is to determine a preferred $\boldsymbol{\beta}$, via training using $(\mathbf{x}_i,y_i)$ pairs, that minimises a suitable loss function; i.e., minimises a suitable measure of the error between the predicted values $\hat{y}_i$ and target values $y_i$. 
One has ``fitted the model to the data'' when a preferred $\boldsymbol \beta$ has been determined.

The standard loss function to optimise is the \emph{mean squared error} (MSE) (i.e., the mean of the squared residuals):

\begin{equation}
 MSE = \frac{1}{n}\sum_{i=1}^n (y_i - f(\mathbf{x}_i,\boldsymbol{\beta)})^2\,.
\end{equation}

In standard multivariable linear modeling, we assume $y = g(\mathbf{x}) + \epsilon $, where $\epsilon$ is a normally distributed random variable representing noise, in which case the solution that optimises the MSE (the least squares solution) also maximises the likelihood of the data \cite{hastie01statisticallearning}.
We optimise $\boldsymbol{\beta}$ using only the training data~--- a separate \emph{test dataset} is used to assess how well the fitted model performs on data (specifically, model inputs) it was not trained on. 
It is common to find that the value of $\boldsymbol{\beta}$ that gives the best model performance on the training set does not give good model performance on the test set. %which is unknown for the model beforehand.\kizito{"which is unknown for the model beforehand" is unclear. If you mean the test set contains data the model was not trained on, this has already been stated.} 

\subsection{Regularization}
Traditional regularization discourages overfitting as follows: during model training, an additional term to the loss function helps find a parameter vector, $\boldsymbol{\beta}$, that ensures $f(\cdot,\boldsymbol{\beta})$ approximates the unknown target function well on unseen data from the same data distribution  \cite{Kukacka2017regsurvey}.
The most common additional terms are the $L_1$ and $L_2$ norms of the parameter vector (denoted $\|\boldsymbol{\beta}\|_1\mbox{ and } \|\boldsymbol{\beta}\|_2$). These terms are typically applied with a coefficient $\theta \in \mathbb{R}$, that regulates how much large weights %\kizito{are weights model parameters? I have clarified in the text. Please correct if I am wrong.}\tillman{You are right, good to clarify.}\kizito{thanks for confirming. Seen and commented out}
(i.e., large model parameters) are penalised, so that the total loss $L$ is%: \kizito{it is incorrect to use a colon after "is"} 
\begin{equation}
  L = \theta \|\boldsymbol{\beta}\|_i + MSE\,,\quad\mbox{for }i \in \{1,2\}.
\end{equation}
Both $L_1$ and $L_2$ norms encourage the weights to be as small as possible. %, resulting in a more smooth function at the end.\kizito{"resulting in a more smooth function at the end" is unclear and not correct.} 
The $L_1$ norm, $\|\boldsymbol{\beta}\|_1 = \sum_{j=1}^m |\boldsymbol{\beta}_j|$, corresponds to a Laplacian prior with 0 mean \cite{williams1995bayesian,plaut86,tibshirani1996L1}. The $L_2$ norm, $\|\boldsymbol{\beta}\|_2 = \sqrt{ \sum_{j=1}^m \boldsymbol{\beta}_j^2}$, corresponds to a Gaussian prior with 0 mean and standard deviation $\theta^{-1}$ \cite{rennie2003l2,Bishop95}.

%%%%%%%%%%%%%%%%%%%%%%%%%%%%%%%%%%%%%

\begin{figure}[t]
  \centering
  \includegraphics[trim={0cm 8.2cm 0cm 10.3cm},clip,scale=0.2]{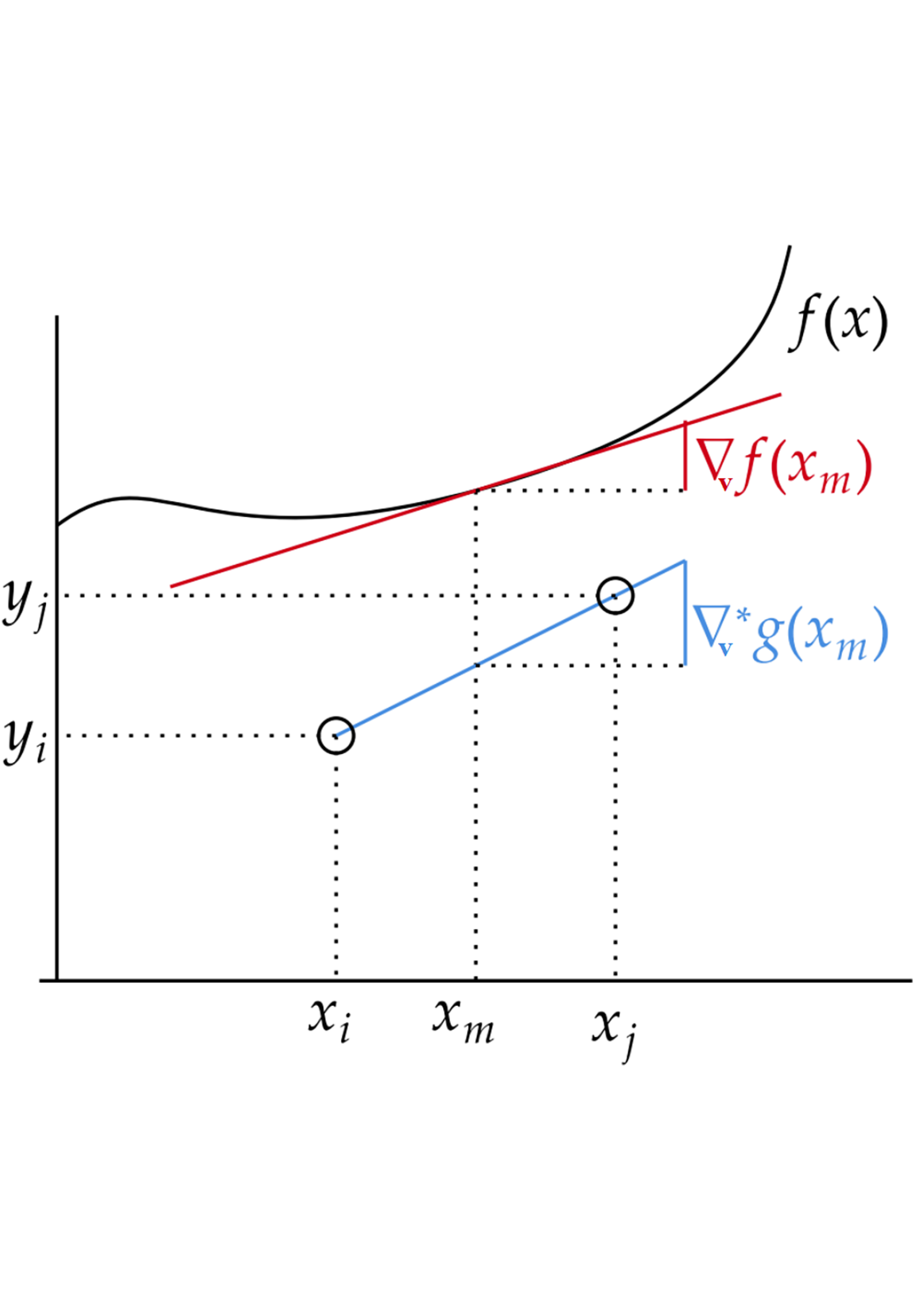}
  \caption{\emph{DLoss} approach in the one-dimensional case for a tuple of pairs $((\mathbf{x}_i, y_i),(\mathbf{x}_j, y_j))$.
  On the horizontal axis we show the inputs $x$, which are scalars in this example. On the vertical axis, we show the scalar regression targets.
  In general, for a pair of points $\mathbf{x}_i$ and $\mathbf{x}_j$, we calculate the \emph{midpoint} $\mathbf{x}_m = (\mathbf{x}_i+\mathbf{x}_j)/2$ and the difference vector $\mathbf{v} = \mathbf{x}_j - \mathbf{x}_i$.
  The red line shows the model derivative $\nabla_{\mathbf{v}}f(\mathbf{x}_m)$ calculated at $\mathbf{x}_m$.
  The blue line shows the estimated derivative $\nabla^*_{\mathbf{v}}g(\mathbf{x}_m)$ calculated at $\mathbf{x}_m$.
  %\kizito{For a future version of the paper, better to use line-type rather than line-colour to distinguish curves in a graph.}
  With \emph{DLoss}, we aim to make the blue and the red lines parallel.
  %\kizito{The notation of the derivatives differs between the caption and the figure. I suggest using the full directional derivative notation in the figure.}\enrico{i will change them now to follow text notation}\enrico{done and commented}
  }
  \label{fig:nablafunct}
\end{figure}

%%%%%%%%%%%%%%%%%%%%%%%%%%%%%%%%%%%%%%%%%%%%%%%%%

\subsection{Derivative Error}\label{subsec_regulmethod}

We propose a new regularization function, \emph{DLoss}, that is defined on the difference between (an estimate of) the model's derivative and the target's derivative (estimated from the training data).
We use the finite-difference method (FDM) to estimate both derivatives \cite{paszynski2021deep,samaniego2020energy}. 
Intuitively, the main idea of \emph{DLoss} is to align the slope of our model $f(\mathbf{x},\boldsymbol{\beta})$ and the slope of the unknown target function $g(\mathbf{x})$, as illustrated in \autoref{fig:nablafunct}.
Since we are dealing with %continuously differentiable 
functions of multi-dimensional data, we use \emph{directional derivatives}.
%In order to align derivatives of $f(\mathbf{x})$ with derivatives of $g(\mathbf{x})$, we need data derivatives. We denote the data derivative $\nabla^*_{\mathbf{v}}g$, which is an estimate of the directional derivative of $g(\mathbf{x})$ in the direction of vector $\mathbf{v}$.

\subsubsection{Data Derivative}

A \emph{data derivative}, denoted $\nabla^{*}_{\mathbf{v}}g$, is an estimate of the directional derivative of $g$ in the direction of vector $\mathbf{v}$. %, the unknown function which describes the data generating process.
When calculating  $\nabla^{*}_{\mathbf{v}}g$ we select a tuple, $((\mathbf{x}_i,y_i),(\mathbf{x}_j,y_j))$, from the training data set.
We calculate the \emph{midpoint} $\mathbf{x}_m = (\mathbf{x}_i+\mathbf{x}_j)/2$ between the points $\mathbf{x}_i$ and $\mathbf{x}_j$, and we obtain the difference vector $\mathbf{v} = \mathbf{x}_j - \mathbf{x}_i$.
Then, the data derivative at $\mathbf{x}_m$ along the direction $\mathbf{v}$ is%: \kizito{it is incorrect to use a colon after "is"} 
\begin{equation} 
\nabla^{*}_{\mathbf{v}}g(\mathbf{x}_m) = \frac{y_j - y_i}{\|\mathbf{v}\|_2}\,.
\end{equation}
In general, one cannot give any guarantees as to how far off the estimated derivative $\nabla^{*}_{\mathbf{v}}g$ may be from the true derivative $\nabla_{\mathbf{v}}g$. 
However, $\nabla^{*}_{\mathbf{v}}g(\mathbf{x}_m)$ is the value of the directional derivative at \emph{some} point between $\mathbf{x}_{i}$ and $\mathbf{x}_{j}$. More formally, consider all points $\mathbf{x}_{\tau}$ of the form 
\begin{equation}
  \mathbf{x}_{\tau} = \tau \mathbf{x}_{i} + (1-\tau)\mathbf{x}_{j}
\end{equation} 
with $0\leq\tau\leq1$.
By the mean value theorem \cite[p 613]{rodhe-et-al-2012-introduction}, if $g$ is continuous over $\{\mathbf{x}_{\tau}|0\leq\tau\leq1\}$ and differentiable over $\{\mathbf{x}_{\tau}|0<\tau<1\}$, there is some $\tau$ such that 
\begin{equation}
 \nabla_{\mathbf{v}}g(\mathbf{x}_{\tau}) = \nabla^{*}_{\mathbf{v}}g(\mathbf{x}_m)\,.
\end{equation}

\subsubsection{Model Derivative}

Consider the function $f(\mathbf{x},\boldsymbol{\beta})$, viewed as a function of only $\mathbf{x}$ with fixed $\boldsymbol{\beta}$, and denote this $f(\mathbf{x})$. We can obtain the value of $\nabla_{\mathbf{v}}f(\mathbf{x})$~--- i.e., the directional derivative of $f$ at point $\mathbf{x}$ in the direction $\mathbf{v}$~--- either analytically or numerically. 
As with target derivatives, we use an FDM approach to approximate $\nabla_{\mathbf{v}}f(\mathbf{x})$ with $\nabla^{\diamondsuit}_{\mathbf{v}}f(\mathbf{x})$, by calculating%: \kizito{it is incorrect to use a colon after "calculating"} 
\begin{equation} \nabla^{\diamondsuit}_{\mathbf{v}}f(\mathbf{x}) = \frac{ f(\mathbf{x}+\varepsilon\tilde{\mathbf{v}})-f(\mathbf{x}-\varepsilon\tilde{\mathbf{v}})}{2\varepsilon}
\end{equation} 
where $\tilde{\mathbf{v}} = \frac{\mathbf{v}}{\|\mathbf{v}\|_2}$ is normalised to unit length and $\varepsilon$ is a small constant scalar. %\kizito{why aren't we using standard notation for unit vectors, $\hat{\mathbf{v}}$? I had changed it but it has been changed back to $\tilde{\mathbf{v}}$.}\enrico{hi K, we havent touched the hat, however we used tilde because hat is used for y prediction}\kizito{seen and commented out}
This approach is more universal, in the sense that one does not need to determine the directional derivative analytically and (using automatic differentiation) it is easy to implement as a loss function. 

\subsubsection{Tuple Selection}

\begin{figure}[t]
  \centering
  \includegraphics[trim={0cm 0cm 0cm 0cm},clip,scale=0.3]{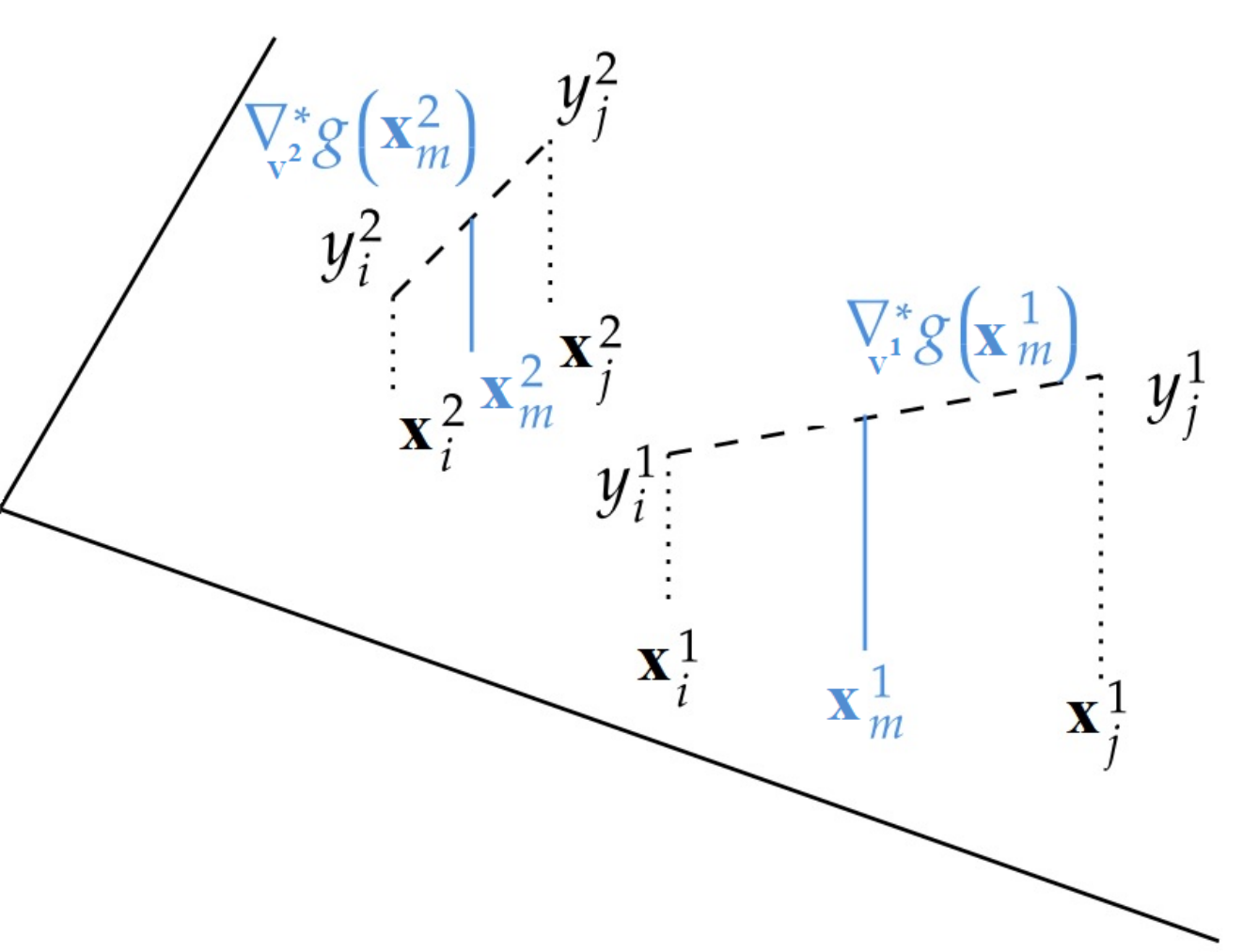}
  \caption{Illustration of data derivatives calculated from two tuples of pairs. 
  For each tuple $((\mathbf{x}^s_i, y^s_i),(\mathbf{x}^s_j, y^s_j))$, where $s\in\{1,2\}$, we calculate the midpoint $\mathbf{x}^s_m$, difference vector $\mathbf{v}^s$, and the corresponding data derivative $\nabla^{*}_{\mathbf{v}^s}g(\mathbf{x}^s_m)$  over a 2-dimensional feature space ($\mathbf{x} \in {\mathbb R}^2$).
%\kizito{as with the other figure, use the full notation (with vector) for directional derivatives that is used in the body of the text. For example, in the figure, use $\nabla^{*}_{\mathbf{v}^2}g(\mathbf{x}^2_m)$.}\kizito{Might be useful, but not necessary, to have a 3rd vertical axis.} \enrico{same here will fix the picture with text notation}\enrico{done and commented}
  }
  \label{fig:tuples}
\end{figure}

%For every training pair $(\mathbf{x}_i, y_i)$, $l$ other pairs $(\mathbf{x}_j, y_j)$ with $j \neq i$ are selected to form tuples that are used to calculate $l$ data gradients.
We used two alternative algorithms for selecting tuples. For each training pair $(\mathbf{x}_i, y_i)$, each algorithm selects $l$ other pairs and, using these pairs, constructs $l$ tuples of the form $((\mathbf{x}_i, y_i), (*, *))$. So, $l$ is a training hyperparameter. These tuples are used to estimate derivatives in section \ref{subsec_optimization}. %\kizito{I think I understand. Please correct my suggested rephrasing if needed.}\enrico{reads ok}\kizito{seen and commented out}
%\kizito{please rewrite -- this is unclear. The figure uses $l$ to index tuples, but here $l$ indicates a number of pairs.}\tillman{changed letter in text above}\kizito{seen and commented out}  
In \autoref{fig:tuples} we sketch an example of multiple tuples in a multi-dimensional space. 

\textbf{Nearest neighbour selection. } 
For each training pair $(\mathbf{x}_i, y_i)$, we select its $l$ nearest neighbours. The pair $(\mathbf{x}_j,y_j)$, where $j\neq i$, is a nearest neighbour if the distance $\|\mathbf{x}_j-\mathbf{x}_i\|_2$ is minimal.
We determine nearest neighbours using the efficient KD-tree algorithm~\cite{nnalgo}. Once a pair is selected as a nearest neighbour, the next nearest neighbour is selected from the remaining pairs in the training set, until $l$ neighbours in total have been selected. 
%If for a pair $p$ a tuple with a selected pair $p_{n1}$ already exists, the next available nearest neighbour pair, e.g. $p_{n2}$ is selected for a tuple and $l$ tuples are still created for $p$.  
%\enrico{please check the above meaning}\kizito{yes, this is confusing}\tillman{I just rewrote, should be clearer now.}\kizito{I think I understand now. Please correct my suggested rephrasing if needed.}\enrico{reads ok}\kizito{seen and commented out}

This approach approximates the derivative based only on local information. 
However, if there is noise in the data (see section \ref{subsec_regressiontasks}) that creates a deviation of  $\epsilon$ from the true value of $g(\mathbf{x}_j)-g(\mathbf{x}_i)$, this will lead to 
$ \epsilon / \|\mathbf{x}_j-\mathbf{x}_i\|_2 $ deviation of the estimate $\nabla_\textbf{v}^* g(\mathbf{x}_m)$, so that small distances between points can amplify noise. 

\textbf{Random selection. } 
For each training pair $(\mathbf{x}_i, y_i)$, we randomly choose $l$ other pairs from the training set. 
%We take $l$ pairs $(\mathbf{x}_j, y_j)$ with $j \neq i$ randomly from the training set. 
%\kizito{As before, I am confused by the use of $l$ here. The reader is tempted to imagine $j$ being an index that runs up to $l$. Please restate with clearer indexing notation that is consistent with the rest of the paper.}
%\tillman{fixed by replacing l earlier, see above}\kizito{Thank you. I think I understand now. Please correct my suggested rephrasing.}\enrico{reads ok}\kizito{seen and commented out}
The idea here is that, for noisy data, there is less influence on $\nabla_\textbf{v}^* g(\mathbf{x}_m)$ since the distances between points are greater. 
The trade-off is that greater distances between tuples lead to lower accuracy of the estimated derivatives.

The random selection has $\mathcal{O}(nl)$ time complexity while the nearest neighbours point selection with the KD-tree has $\mathcal{O}(nl + n \log(n))$, which may make random selection preferable for very large datasets.%\kizito{The ``which may play a role'' comment is too cryptic, Explicitly state/speculate what role might be played.} \tillman{done}\kizito{seen and commented out}

\subsubsection{Optimization}
\label{subsec_optimization}
The \emph{derivative difference} $dd(\mathbf{x})$ is the estimate of the misalignment between model and target derivatives: 
\begin{equation}
  dd(\mathbf{x}) = \nabla^{\diamondsuit}_{\mathbf{v}}f(\mathbf{x}) - \nabla^{*}_{\mathbf{v}}g(\mathbf{x})\,.
\end{equation}

Let $\mathcal S$ be an index set for all tuples generated by one of the tuple selection approaches above. For the tuple with index $s\in\mathcal S$, we calculate the midpoint, $\mathbf{x}_m^{s}$, and the derivatives' estimates, $\nabla^{\diamondsuit}_{\mathbf{v}^s}f(\mathbf{x}_m^{s})$ and $\nabla^{*}_{\mathbf{v}^s}g(\mathbf{x}_m^{s})$. 
The \emph{DLoss} aggregates the $dd$ values as the mean squared derivative-difference:%\kizito{Because I don't understand the use of $l$ in the selection descriptions, I cannot judge if the following expression makes sense}\tillman{$l$ is the hyperparameter for \# of tuples, should be clear now.}\kizito{I think I understand the selection algorithms now. I don't think the index $j$ below works, since different ``$i$''s will have different ``$j$''s for, say, nearest neighbours selection. Also, $j\neq i$ needs to be enforced. Please correct my rephrasing in terms of $\mathcal S$ if needed.}\enrico{reads ok}\kizito{seen and commented out}
%\begin{equation}
%  DLoss = \frac{1}{ln} \sum_{i=1,j=1}^{n,l} dd(\mathbf{x}_m^{i,j})^2
%\end{equation}

\begin{equation}
  DLoss = \frac{1}{|\mathcal S|} \sum_{s\in\mathcal S} dd(\mathbf{x}_m^{s})^2
\end{equation}

where $|\mathcal S|$ is the cardinality of $\mathcal S$. The total loss $L$ for training the model is%: \kizito{it is incorrect to use a colon after "is".}
\begin{equation}
  L = \theta_{D} \cdot DLoss + MSE
\end{equation}
where $\theta_{D}$ is a weighting coefficient. 
This loss is minimised with gradient descent. 

%%%%%%%%%%%%%%%%%%%%%%%%%%%%%%%%%%%%%

\section{Experiments}\label{sec:experiments}

%%%%%%%%%%%%%%%%%%%%%%%%%%%%%%%%%%%%%

\textbf{Setup} 

In our experiments, the regression model is a simple feed-forward neural network with a single hidden layer (consisting of $L_H$ neurons)
%\kizito{$L-h$ uses a subscript that is also an index in the summation expression. So, changed to $L_H$.}
and ReLU activation. %\kizito{We should explicitly state how many hidden neurons we used?}\tillman{it's stated below. This will change and we'll use multiple values in future experiments, so probably better stated separately.}\kizito{Thank you; I added the definition of $L_h$ here but had not seen the stated value below. I'll comment out our comments} 
We use ReLU activation because of its popularity in modern neural networks~\cite{goodfellow2016deep}.
The network calculates the function
$\hat{y} = f(\mathbf{x},\boldsymbol{\beta}) =
\sum^{L_H}_{h=1} w_{o,h}  r(\mathbf{w}_{h} \cdot \mathbf{x} + b_{h}) + b_{o}$, where $\mathbf{w}_{h}$ and ${b}_{h}$ are the  weight vector and bias value of hidden neuron $h$, respectively, and ``$\cdot$'' denotes the scalar product.
The weight between hidden neuron $h$ and the output neuron is $w_{o,h}$, while $b_{o}$ is the bias of the output neuron. All of these weights and biases make up the vector $\boldsymbol{\beta}$.  
The rectified linear function (ReLU) $r:\mathbb{R}\rightarrow\mathbb{R}$, defined as $r(x) = max(x,0)$, was first introduced for neural networks (to the best our knowledge) by \cite{fukushima1975relu}. 
%\kizito{How are $\boldsymbol{\beta}$, $\mathbf{b}_i$ and $b_0$ related?} \enrico{$\boldsymbol{\beta} = (\mathbf{b}_i,b_o,\mathbf{w}_i,w_o)$}\kizito{be more careful; $i$ takes on several values, so there are several $\mathbf{b}_i$ and $\mathbf{w}_i$.}\enrico{i in this case refers to input and ''o'' for output}
%\kizito{That does not make sense -- the model parameters are not model inputs. Besides, there is a summation...indexed by $i$? BTW, the summation does not have a range stated -- it should.}
%\enrico{i shall rephrase, its written in the text that i are the hidden neurons}\tillman{Kizito, the i and o are just indeces, not values. Enrico, there is an inconsistency here. the weights are between layers, name then either i,h or h,,o}\enrico{done}\tillman{not what I had in mind, but that works}

In our experiments, we use the Adam optimiser \cite{kingma2015adam}.
We compare the effects of using ${DLoss}$ in combination with MSE, against the $L_2$ and Dropout ($DO$) regularization methods. 
We also compare the different variants of tuple selection - Random ($DL_{RND}$) and Nearest neighbours ($DL_{NN}$). 

\paragraph{Model parameters} Our models have an input layer with a number of neurons determined by the dimensionality of the datasets, a single hidden layer with $L_H = 64$ hidden neurons, and a single linear output neuron. 
We set $\varepsilon$ to $0.001$ for calculating the model derivative. 
%and the learning rate $\lambda$ as explained in the hyper-parameters section. 
We use 5-fold cross validation. 

%%%%%%%%%%%%%%%%%%%%%%%%%%%%%%%%%%%%%

\textbf{Data}
In order to observe the effect of our approach in different contexts, we conduct experiments using real data and synthetic, noiseless data. 
Five commonly available real datasets have been selected: Wine, Cancer, Modechoice, Anes96, Diabetes; further information is given in \autoref{tab:data-table}.
For noiseless data, we created synthetic datasets starting using data generating functions available from the Scikit-Learn library, where we have selected F1, Regression1, Regression10, Sparse-uncorrelated, Swiss Roll. 
For these, the training data range of the input $\mathbf{x}$ is in the range $[0,1]$ with a fixed $2500$ uniformly distributed points and $1$, $3$, or $10$ features.
\autoref{tab:data-table} shows the main features of the datasets.

\begin{table}[tb]
 \begin{center}
 \begin{small}
 \begin{sc}
 \begin{tabular}{llrrcccc}
 \toprule
 Name & Type & Features & Points & Library & Description & Year\\
 
  \midrule
  anes96 & Real & 5 & 944 & Statsmodel & US National Election Survey & 1996\\
  cancer & Real & 1 & 301 & Statsmodel & Breast Cancer Observation & 2007 \\
  diabetes & Real & 10 & 442 & Scikit - NSCU & Diabetes Observation & 2004\\
  modechoice & Real & 6 & 840 & Statsmodel & Travel Choice & 1987 \\
  wine & Real & 11 & 1599 & UCI & Wine Quality & 1988 \\
  \cline{2-7} 
  f1 & Synth & 10 & 2500 & Scikit & \texttt{\textup{make\_friedman1}} \\
regression1 & Synth & 1 & 2500 & Scikit & \texttt{\textup{make\_regression}} \\
  regression10 & Synth & 10 & 2500 & Scikit & \texttt{\textup{make\_regression}}\\
  sparse uncorr & Synth & 10 & 2500 & Scikit & \texttt{\textup{make\_sparse\_uncorrelated}}\\
  swiss roll & Synth & 3 & 2500 & Scikit & \texttt{\textup{make\_swiss\_roll}} \\

  \bottomrule
  \end{tabular}
  \end{sc}
  \end{small}
  \end{center}
  \caption{Datasets used in our experiments. All datasets have a single real-valued output. 
  The number of inputs is listed in column \textsc{Features}. 
  \textsc{Library} describes the Python package from which the dataset is available.
  The synthetic datasets (\textsc{Type: Synth}) were generated with the Python library \emph{Scikit-Learn}, version 1.4.1, using the methods from the package $sklearn.datasets$. 
  In the generation of all datasets, no noise has been added.
  Further details of the data generating functions can be found in the library documentation available at \texttt{\textup{https://scikit-learn.org/stable/datasets/}}.
  }
  \label{tab:data-table} 
\end{table}
%%%%%%%%%%%%%%%%%%%%%%%%%%%%%%%%%%%%%

\textbf{Training Hyper-parameters}
We ran a grid search over the following hyper-parameters: learning rate $\lambda = 0.03,0.01,0.003,0.001$, $L_2$ weight $\theta = 10^{[-3,-4,-5,-6,-7]}$, \emph{DLoss} weight $\theta_{D} = 10^{[-3,-4,-5,-6,-7]}$, and Dropout probability $p = [0.05,0.1,0.2,0.4,0.8]$.
Regularizations $L_2$, $DO$, and \emph{DLoss}, are not combined in our experiments. 
We train our models for 250 epochs and use full batch learning. 

\textbf{Metrics}

We report the first epoch at which the best accuracy result is achieved ($ep$) and the time ($t$) for the full training (250 epochs).
The metrics reported are calculated over the 5 cross-validation folds with mean and standard deviation: 
\begin{itemize}
 \item \textbf{$MSE_{train}$}: the average of the best MSE value per fold on the training set over all epochs;
 \item \textbf{$\sigma_{{MSE}_{train}}$}: standard deviation of the best MSE value per fold on the training set over all epochs;
 \item \textbf{$MSE_{val}$}: average of the best MSE value per fold on the validation set over all epochs;
 \item \textbf{$\sigma_{{MSE}_{val}}$}: standard deviation of the best MSE value per fold on the validation set over all epochs;
 \item \textbf{$ep$}: average of the epoch at which the minimum $MSE_{val}$ is reached per epoch;
 \item \textbf{$t$}: average time in seconds to complete a single training per single fold.
\end{itemize}

%%%%%%%%%%%%%%%%%%%%%%%%%%%%%%%%%%%%%%%%%%%%%%%

\section{Results}\label{sec:results}

In \autoref{tabresults}, we show the results of our approach (best of $DL_{RND}$ and $DL_{NN}$ algorithms) against $STD$ training, $STD+L_2$ and $STD+DO$. 
\emph{DLoss} method shows lower generalisation errors $MSE_{val}$ on both real and synthetic datasets.
Our approach seems more effective with real data.
The lower generalisation error achieved is also displayed in the learning curves reported for the experiments using real datasets in \autoref{fig:learn-real} and for synthetic in \autoref{fig:learn-syn}.
The computation time for \emph{DLoss} is generally higher than the $STD$ models (see $t$ values in \autoref{tabresults}).
In \autoref{tabrank} we report the ranking of $MSE_{val}$ of the different regularization methods per dataset.
Overall, we observe $DL_{NN}$ being ranked first, achieving on average better results than, in worsening rank, $L_2$, $DL_{RND}$, $DO$ and $STD$ models. 

\begin{table}
  \begin{center}
  \begin{small}
  \begin{sc}
  \npdecimalsign{.}
  \nprounddigits{5}
  \begin{tabular}{ll n{1}{5} n{1}{5} n{1}{5} n{1}{5} rr}
  
  \toprule
  Dataset & Method & \multicolumn{1}{c}{$MSE_{{train}}$} & \multicolumn{1}{c}{$\sigma_{{train}}$} & \multicolumn{1}{c}{$MSE_{{val}}$} & \multicolumn{1}{c}{$\sigma_{{val}}$} & $ep$ & $t$ \\
  
  \midrule
\multirow{5}{*}{anes96}  & STD  & 0.177330 & 0.019518 & 0.60008727 & 0.11239327 & 15 & 2.6  \\
 & STD+$L_2$ & 0.30780832 & 0.02352047 & 0.57475274 & 0.09235593 & 41 & 7.7  \\
 & STD+DO & 0.49298516 & 0.03259643 & 0.58150342 & 0.13898003 & 53 & 2.6\\
 & $DL_{RND}$ & 0.19896433 & 0.01540072 & 0.56350781 & 0.1194641 & 12 & 5.5\\
 & $DL_{NN}$ & 0.4295374 & 0.03931913 & {\npboldmath}0.56324297 & 0.09467875 & 53 & 11.8\\ \cline{2-8}
\multirow{5}{*}{cancer}  & STD  & 0.29283602 & 0.01398396 & 0.24764358 & 0.24436089 & 17 & 2.2  \\
 & STD+$L_2$ & 0.30778304 & 0.03125095 & {\npboldmath}0.21662021 & 0.21220799 & 98 & 2.2  \\
 & STD+DO & 0.29449674 & 0.01644076 & 0.23368603 & 0.20838813 & 105 & 7.8  \\
 & $DL_{RND}$ & 0.29207851 & 0.02132386 & 0.22473886 & 0.18396828 & 84 & 3.9  \\
 & $DL_{NN}$ & 0.30993664 & 0.03188799 & 0.21857888 & 0.16864872 & 77 & 3.7  \\ \cline{2-8}
\multirow{5}{*}{diabetes} & STD  & 0.33635206 & 0.05040892 & 0.50646346 & 0.17958295 & 113 & 2.3 \\
 & STD+$L_2$ & 0.14309196 & 0.04330189 & 0.45751889 & 0.11661063 & 43 & 5.3  \\
 & STD+DO & 0.38423959 & 0.01592457 & 0.47632056 & 0.131764 & 133 & 13.7  \\
 & $DL_{RND}$ & 0.11744604 & 0.02434119 & 0.45636614 & 0.17695314 & 32 & 4.3  \\
 & $DL_{NN}$ & 0.00893422 & 0.00529294 & {\npboldmath}0.4553714 & 0.23124737 & 23 & 5.9  \\ \cline{2-8}
\multirow{5}{*}{modechoice} & STD  & 0.1179477 & 0.01804242 & 0.63944455 & 0.14480025 & 131 & 2.6\\
 & STD+$L_2$ & 0.14903403 & 0.03706822 & 0.58849685 & 0.05570832 & 179 & 7.5  \\
 & STD+DO & 0.44150581 & 0.01733288 & 0.53832477 & 0.07132203 & 186 & 15.4  \\
 & $DL_{RND}$ & 0.13025415 & 0.01307876 & 0.59405038 & 0.08625778 & 99 & 9.1  \\
 & $DL_{NN}$ & 0.14156311 & 0.01416364 & {\npboldmath}0.53603314 & 0.16874024 & 147 & 8.8  \\ \cline{2-8}
\multirow{5}{*}{wine} & STD  & 0.09746584 & 0.01039641 & 0.59575467 & 0.11823012 & 17 & 3.1 \\
 & STD+$L_2$ & 0.08724606 & 0.00577489 & 0.55471867 & 0.13983683 & 31 & 43.3  \\
 & STD+DO & 0.44190202 & 0.01508975 & 0.5916705 & 0.16977146 & 117 & 93.7  \\
 & $DL_{RND}$ & 0.47699501 & 0.02936241 & {\npboldmath}0.5496235 & 0.23304432 & 167 & 15.1  \\
 & $DL_{NN}$ & 0.11576753 & 0.00938199 & 0.56862197 & 0.13514901 & 36 & 16.3 \\ \cline{2-8}
\multirow{5}{*}{f1} & STD  & 0.00143714 & 0.00022515 & 0.00854035 & 0.08697299 & 241 & 3.7  \\
 & STD+$L_2$ & 0.00136922 & 0.00013204 & {\npboldmath}0.0063039 & 0.08154564 & 248 & 45.8 \\
 & STD+DO & 0.01164792 & 0.00038945 & 0.00673393 & 0.08439192 & 242 & 3.5  \\
 & $DL_{RND}$ & 0.00138717 & 0.00032696 & 0.00688173 & 0.05618682 & 245 & 10.2  \\
 & $DL_{NN}$ & 0.00137461 & 0.00014238 & 0.00655821 & 0.00045686 & 250 & 23.2  \\ \cline{2-8}
\multirow{5}{*}{regression1}  & STD  & 7.14e-6 & 4.03e-6 & 7.64e-6& 0.02236516 & 250 & 3.6\\
 & STD+$L_2$ & 4.93e-6 & 2.3e-6  & 4.84e-6& 0.07453147 & 250 & 44.6\\
 & STD+DO & 0.00274638 & 0.00014872 & 1.459e-5 & 0.0344513 & 217 & 3.5 \\
 & $DL_{RND}$ & 5.26e-6 & 2.93e-6 & 5.86e-6& 0.03172563 & 249 & 10.1\\
 & $DL_{NN}$ & 4.15e-6 & 2.66e-6 & {\npboldmath}4.19e-6& 0.08469988 & 250 & 23.8\\ \cline{2-8}
\multirow{5}{*}{regression10} & STD  & 0.00012851 & 3.658e-5 & 0.00042876 & 0.13315077 & 250 & 3.7  \\
 & STD+$L_2$ & 2.228e-5 & 8.8e-7 & {\npboldmath}2.652e-5 & 0.07846928 & 250 & 57.7\\
 & STD+DO & 0.00451834 & 0.00022454 & 0.0001753 & 0.13754859 & 235 & 3.5\\
 & $DL_{RND}$ & 0.000138  & 2.24e-5 & 0.00039169 & 0.08767906 & 249 & 24.4\\
 & $DL_{NN}$ & 0.0001294 & 2.35e-5 & 0.00037651 & 9.917e-5 & 250 & 9.7 \\ \cline{2-8}
\multirow{5}{*}{sparse uncorr} & STD  & 0.05514032 & 0.0030176 & 0.08518794 & 0.06130903 & 174 & 3.7 \\
 & STD+$L_2$ & 0.02399948 & 0.00171738 & 0.08620409 & 0.09918734 & 42 & 3.8\\
 & STD+DO & 0.17109021 & 0.00405457 & {\npboldmath}0.07938602 & 0.04937745 & 197 & 3.5\\
 & $DL_{RND}$ & 0.0381035 & 0.0025631 & 0.08294402 & 0.14912423 & 45 & 10.0\\
 & $DL_{NN}$ & 0.02399658 & 0.00205902 & 0.08209252 & 0.07535213 & 44 & 23.8 \\ \cline{2-8}
\multirow{5}{*}{swiss roll} & STD  & 0.0001703 & 3.736e-5 & 0.00023135 & 0.10705988 & 249 & 3.7\\
 & STD+$L_2$ & 0.00013068 & 4.676e-5 & 0.00016883 & 0.05157496 & 248 & 39.8\\
 & STD+DO & 0.01236847 & 0.00051966 & 0.00108743 & 0.03987374 & 210 & 3.5\\
 & $DL_{RND}$ & 0.00013275 & 2.252e-5 & 0.00019632 & 0.03456687 & 245 & 23.8\\
 & $DL_{NN}$ & 0.00012357 & 2.436e-5 & {\npboldmath}0.00016706 & 2.355e-5 & 239 & 9.5\\ 
 
 \bottomrule
 \end{tabular}
 \npnoround
 \end{sc}
 \end{small}
 \end{center}

 \caption{$Data$, $Method$ and $Metrics$ of experiments as mean of best results in 5-Fold cross validation over 250 epochs. Neural network of multiple inputs and single output. Network structure of 64 hidden neurons and ReLU activation function. Comparison of $STD$, $STD+L_2$, $STD+DO$ vs $DL_{RND}$ random selection and vs $DL_{NN}$ nearest neighbour selection. 
 In bold are highlighted the best results per datasets. 
 }
 \label{tabresults}
\end{table}

%%%%%%%%%%%%%%%%%%%%%%%%%%%%%%%%%%%%%

\subsection{Discussion}\label{sec:discussion}

In \autoref{tab:bestparam-table}, we report the parameters leading to the best $MSE_{val}$. 
We notice that the learning rate $\lambda$ for synthetic data is on average higher than for real data, with exception for Sparse Uncorr $DL_{RND}$ selection algorithm.
We used 1 and 3 as possible number of tuple per data-points.
%\kizito{I don't understand what ``invariant'' means here, or what ``....an equal amount of numbers'' means. Please rewrite this sentence.}.
%\enrico{basically i wanted to say that 1 is selected as many times as 3 and we dont see any pattern that would suggest the tuple selection method to be biased in any way toward 1 or 3. }\kizito{I still don't understand. If the number of tuples per point is $l$, what is the meaning of ``number of times of selecting $l$ tuples''?}\enrico{l=1 is selected as many times as l=3 which is 10 as per table 5}\enrico{i will remove all this part to avoid confusin}
The weight $\theta_{D}$ associated with the \emph{DLoss} is on average higher for synthetic data than for real dataset.
Moreover, we notice that $\theta_{D}$ for real data is on average higher for $DL_{NN}$, while for synthetic data, $\theta_{D}$ is higher for $DL_{RND}$. 

\paragraph{Trade-offs and optimisations}
The current finite-difference calculation of the model derivative requires two additional forward-passes per tuple when calculating the derivative, one for $f(\mathbf{x}+\varepsilon\tilde{\mathbf{v}})$ and one for $f(\mathbf{x}-\varepsilon\tilde{\mathbf{v}})$.
The increase in training time could only partly be offset by faster convergence if early stopping were used. 
It may be possible to reduce the cost by not determining the model derivative at the midpoint but reusing data points instead or by using analytical gradient calculations, which would be model dependent. 
There may also be opportunities for algorithmic improvement.
Neither of these options were used in this study. 

%%%%%%%%%%%%%%%%%%%%%%%%%%%%%%%%%%%%%

\subsection{Significance Tests} 

We list the median values, $Median_\Delta$, of the paired difference for $DL$ $MSE_{val}$ and $STD$ $MSE_{val}$ with significance test results in \autoref{tabtests}.
In order to test whether the observed differences in $MSE_{val}$ values are statistically significant, we apply a Wilcoxon Signed Rank Test and a paired t-test.
The common approach would suggest to use t-test, which tests for different averages.
This, however, assumes normality of the distribution of the differences.
We therefore test for normality with Shapiro-Wilk test.
We observe that not in all cases normality of the distribution of the differences is true.
We therefore run the Wilcoxon Signed Rank Test, which tests for different medians.

We consider the best $MSE_{val}$ results as data-points.
We divide the results into real and synthetic data, with 5 dataset for each category.
We use four groups of results according to the learning methods $STD$, $STD+L_2$, $DL_{RND}$, $DL_{NN}$ and compare each $STD$ with each $DL$ group.
The data-points used are the $MSE_{val}$ results for each of the 5 folds in each dataset/method combination, leading to 25 data-points.

The observed improvement of $Median_\Delta$ for $DL_{RND}$ over $STD$ method is statistically significant for real data and synthetic datasets.
The improvement we observe are not always statically significant according to \autoref{tabtests}.
\emph{DLoss} produces better, and in three cases also significant, results than $STD$, while not against $L_2$ or $DO$.
We observe that the $Median_\Delta$ is not positive for all pairs.
For those results, the Wilcoxon Signed Rank Test is not significant, therefore we cannot significantly confirm that \emph{DLoss} does not produce improvement over $L_2$.
We ran the Shapiro-Wilk test, which showed a significant difference from normal distribution in all but three cases, out of which the t-test did not show a significant difference between $STD$ and $DL$ group.

\begin{table}[tb]
  \begin{center}
  \begin{small}
  \begin{sc}
  \begin{tabular}{ccc|llll}
  
  \toprule
  Dataset & Std Group & DLoss Group & $Median_{\Delta}$ & Wilcoxon $p$ & Shapiro-Wilk $p$ & t-test $p$\\
  
  \midrule
  \multirow{6}{*}{Real} & STD & $DL_{RND}$ & 1.3$\times10^{-2}$ & $0.035^{*}$ & $0.046^{*}$ & $0.057$ \\
 &STD & $DL_{NN}$ & 4.2$\times10^{-2}$ & 0.078 & 0.37 & $0.044^{*}$\\
 &STD+$L_2$& $DL_{RND}$ & -8.6$\times10^{-3}$ & 0.53 & 0.75 & 0.48 \\
 &STD+$L_2$& $DL_{NN}$ & -4.1$\times10^{-4}$ & 0.32 & $0.022^{*}$ & 0.28 \\
 &STD+DO& $DL_{RND}$ & 1.8$\times10^{-2}$ & 0.19 & $0.017^{*}$ & 0.38 \\
 &STD+DO& $DL_{NN}$ & 4.5$\times10^{-3}$ & 0.25 & 0.56 & 0.21 \\
  
  \hline
  \multirow{6}{*}{Synthetic} & STD & $DL_{RND}$ & 3.9$\times10^{-5}$ & $0.005^{**}$ & 3.6$\times10^{-6***}$ & 0.157\\
 &STD & $DL_{NN}$ & 1.8$\times10^{-5}$ & $0.011^{*}$ & 2.3$\times10^{-5***}$ & 0.121 \\
 &STD+$L_2$& $DL_{RND}$ & -2.9$\times10^{-5}$ & 0.85 & 6.7$\times10^{-6***}$ & 0.241\\
 &STD+$L_2$& $DL_{NN}$ & -2.7$\times10^{-5}$ & 0.77 & 5.6$\times10^{-7***}$ & 0.144\\
 &STD+DO& $DL_{RND}$ & 6.9$\times10^{-6}$ & 0.42 & 7.3$\times10^{-6***}$ & 0.86 \\
 &STD+DO& $DL_{NN}$ & 1.2$\times10^{-5}$ & 0.26 & 2.2$\times10^{-6***}$ & 0.49 \\
  
  \bottomrule
  \end{tabular}
 
  \end{sc}
  \end{small}
  \end{center}

  \caption{The results of significance tests for differences of $MSE_{val}$ between $STD$ and \emph{DLoss} results. 
  We run the tests on the difference of the standard learning method group - $STD$ group - and the group with \emph{DLoss} - $DL$ group.
  For both $STD$ and $DL$ training we have 2 group, therefore we run all combinations, giving us 6 results, for real data and synthetic.
  We compare each of the 5-fold in the cross validation for each of the 5 datasets, thus 25 samples per test.
  $Median_{\Delta}$ is the median of pairwise difference between the $MSE_{val}$ values of the groups.
  We applied the Wilcoxon Signed Ranked Test, a non-parametric test for significance of differences between the medians. 
  The low $p$-values of the signed rank test indicate that the $STD$ groups' $MSE_{val}$ is significantly greater than the and $DL$ groups' (indicated with an asterisk *). 
  The $^{*},^{**},^{***}$ indicate the significance at 5\%, 1\% and 0.1\% %\kizito{0.1\%?}\enrico{fixed}\kizito{seen and commented out} 
  confidence, respectively.
  We also applied a paired t-test. 
  However, the t-test is only strictly valid for normally distributed data, and the Shapiro-Wilk test indicates that the distribution is significantly different from a normal distributions.  
  }
  \label{tabtests}
\end{table}

\begin{table}[tb]
  \begin{center}
  \begin{small}
  \begin{sc}
  \begin{tabular}{llcccc}
  
  \toprule
\multirow{2}{*}{Dataset} & & \multicolumn{2}{c}{STD GROUP}&  \multicolumn{2}{c}{DLoss GROUP}\\
& STD & STD+$L_2$& STD+DO& $DL_{RND}$& $DL_{NN}$\\
  
  \midrule
  anes96  & 5 & 3& 4& 2& 1\\
  cancer & 5 & 1& 4& 3& 2\\
  diabetes  & 5 & 3& 4& 2& 1\\
  modechoice & 5 & 3& 2& 4& 1\\
  wine  & 5& 2& 4& 1& 3\\
  f1 & 5& 1& 3& 4& 2\\
  regression1 & 4& 2& 5& 3& 1 \\
  regression10& 5& 1& 2& 4& 3 \\
  sparse uncorr & 4& 5& 1& 3& 2 \\
  swiss roll & 4& 2& 5& 3& 1 \\
  \hline
  $Avg$& 4.7 & 2.3& 3.4& 2.9& 1.7 \\ 
  \bottomrule
  
  \end{tabular}
  \end{sc}
  \end{small}
  \end{center}
  \caption{The ranks of methods $STD$, $STD+L_2$, $STD+DO$, $DL_{RND}$ and $DL_{NN}$ according to $MSE_{val}$ by $Dataset$ and their averages.}
  \label{tabrank}  
\end{table}

%%%%%%%%%%%%%%%%%%%%%%%%%%%%%%%%%%%%%

\section{Conclusion}\label{sec:conclusion}
We propose a new regularization method, \emph{DLoss}, which aims to reduce the difference between model derivatives and  target function derivatives by using differential information inferred from training data.
The \emph{DLoss} method is tested on ten regression datasets.
We propose two versions of the method with different tuple-selection algorithms~--- random and nearest neighbour selection.
We experiment on a neural network with a single hidden layer, multiple inputs, and a single output. 
We used the MSE error function on validation sets to compare the performance of different regularization methods. 
We observed an overall generalisation improvement using \emph{DLoss} with nearest neighbour selection ($DL_{NN}$). 
With this method, the validation error is lower and, in some cases, convergence is reached with fewer epochs of training, compared to $L_2$ and Dropout regularization.
Depending on the application, the improved generalisation may come with additional computational cost; in this regard, further optimisation may still be possible. 
Future work will include tests with more models and more datasets, as well as the formulation of \emph{DLoss} for classification problems.
Moreover, higher dimensional datasets with more data points and more model architectures are needed in order to determine whether the benefits of using the \emph{DLoss} method generalise further. 

%%%%%%%%%%%%%%%%%%%%%%%%%%%%%%%%%%%%%

\bibliographystyle{apalike} 
\bibliography{references} 

\begin{thebibliography}{}

\bibitem[Avrutskiy, 2017]{Avrutskiy2017}
Avrutskiy, V.~I. (2017).
\newblock Backpropagation generalized for output derivatives.
\newblock {\em CoRR}, abs/1712.04185.

\bibitem[Avrutskiy, 2021]{Avrutskiy2021}
Avrutskiy, V.~I. (2021).
\newblock Enhancing function approximation abilities of neural networks by training derivatives.
\newblock {\em IEEE Transactions on Neural Networks and Learning Systems}, 32(2):916--924.

\bibitem[Baldi and Sadowski, 2013]{baldi2013understanding}
Baldi, P. and Sadowski, P.~J. (2013).
\newblock Understanding dropout.
\newblock {\em Advances in neural information processing systems}, 26.

\bibitem[Beck et~al., 2023]{beck2023overview}
Beck, C., Hutzenthaler, M., Jentzen, A., and Kuckuck, B. (2023).
\newblock An overview on deep learning-based approximation methods for partial differential equations.
\newblock {\em Discrete \& Continuous Dynamical Systems-Series B}, 28(6).

\bibitem[Bishop, 1995]{Bishop95}
Bishop, C.~M. (1995).
\newblock {\em Neural networks for pattern recognition}.
\newblock Clarendon Press, Oxford.

\bibitem[Chen et~al., 2018]{NEURIPS2018_Chen}
Chen, R. T.~Q., Rubanova, Y., Bettencourt, J., and Duvenaud, D.~K. (2018).
\newblock Neural ordinary differential equations.
\newblock In Bengio, S., Wallach, H., Larochelle, H., Grauman, K., Cesa-Bianchi, N., and Garnett, R., editors, {\em Advances in Neural Information Processing Systems}, volume~31. Curran Associates, Inc.

\bibitem[Cuomo et~al., 2022]{cuomo2022scientific}
Cuomo, S., Di~Cola, V.~S., Giampaolo, F., Rozza, G., Raissi, M., and Piccialli, F. (2022).
\newblock Scientific machine learning through physics--informed neural networks: Where we are and what’s next.
\newblock {\em Journal of Scientific Computing}, 92(3):88.

\bibitem[Fukushima, 1975]{fukushima1975relu}
Fukushima, K. (1975).
\newblock Cognitron: A self-organizing multilayer neural network.
\newblock {\em Biological Cybernetics}, 20:121--136.

\bibitem[Goodfellow et~al., 2016]{goodfellow2016deep}
Goodfellow, I., Bengio, Y., and Courville, A. (2016).
\newblock {\em Deep learning}.
\newblock MIT press.

\bibitem[Hastie et~al., 2001]{hastie01statisticallearning}
Hastie, T., Tibshirani, R., and Friedman, J. (2001).
\newblock {\em The Elements of Statistical Learning}.
\newblock Springer Series in Statistics. Springer New York Inc., New York, NY, USA.

\bibitem[Hinton et~al., 2012]{hinton2012L2}
Hinton, G.~E., Srivastava, N., Krizhevsky, A., Sutskever, I., and Salakhutdinov, R.~R. (2012).
\newblock Improving neural networks by preventing co-adaptation of feature detectors.

\bibitem[Hornik et~al., 1990]{hornik-et-al-1990-universal}
Hornik, K., Stinchcombe, M., and White, H. (1990).
\newblock Universal approximation of an unknown mapping and its derivatives using multilayer feedforward networks.
\newblock {\em Neural Networks}, 3(5):551--560.

\bibitem[Kidger, 2022]{Kidger22}
Kidger, P. (2022).
\newblock On neural differential equations.
\newblock {\em CoRR}, abs/2202.02435.

\bibitem[Kingma and Ba, 2015]{kingma2015adam}
Kingma, D. and Ba, J. (2015).
\newblock Adam: A method for stochastic optimization.
\newblock In {\em International Conference on Learning Representations (ICLR)}, San Diega, CA, USA.

\bibitem[Kukacka et~al., 2017]{Kukacka2017regsurvey}
Kukacka, J., Golkov, V., and Cremers, D. (2017).
\newblock Regularization for deep learning: {A} taxonomy.
\newblock {\em CoRR}, abs/1710.10686.

\bibitem[Lee and Kang, 1990]{lee1990neural}
Lee, H. and Kang, I.~S. (1990).
\newblock Neural algorithm for solving differential equations.
\newblock {\em Journal of Computational Physics}, 91(1):110--131.

\bibitem[Malek and Beidokhti, 2006]{malek2006numerical}
Malek, A. and Beidokhti, R.~S. (2006).
\newblock Numerical solution for high order differential equations using a hybrid neural network—optimization method.
\newblock {\em Applied Mathematics and Computation}, 183(1):260--271.

\bibitem[Parisi et~al., 2003]{parisi2003solving}
Parisi, D.~R., Mariani, M.~C., and Laborde, M.~A. (2003).
\newblock Solving differential equations with unsupervised neural networks.
\newblock {\em Chemical Engineering and Processing: Process Intensification}, 42(8-9):715--721.

\bibitem[Paszy{\'n}ski et~al., 2021]{paszynski2021deep}
Paszy{\'n}ski, M., Grzeszczuk, R., Pardo, D., and Demkowicz, L. (2021).
\newblock Deep learning driven self-adaptive hp finite element method.
\newblock In {\em International Conference on Computational Science}, pages 114--121. Springer.

\bibitem[Plaut et~al., 1986]{plaut86}
Plaut, D.~C., Nowlan, S.~J., and Hinton, G.~E. (1986).
\newblock Experiments on learning by back propagation.
\newblock Technical report, Carnegie-Mellon University.

\bibitem[Raissi et~al., 2019]{raissi2019physics}
Raissi, M., Perdikaris, P., and Karniadakis, G.~E. (2019).
\newblock Physics-informed neural networks: A deep learning framework for solving forward and inverse problems involving nonlinear partial differential equations.
\newblock {\em Journal of Computational physics}, 378:686--707.

\bibitem[Rennie, 2003]{rennie2003l2}
Rennie, J. (2003).
\newblock On l2-norm regularization and the {G}aussian prior.

\bibitem[Rohde et~al., 2012]{rodhe-et-al-2012-introduction}
Rohde, U.~L., Jain, G.~C., Poddar, A.~K., and Ghosh, A.~K. (2012).
\newblock {\em Introduction to Differential Calculus}.
\newblock Wiley.

\bibitem[Samaniego et~al., 2020]{samaniego2020energy}
Samaniego, E., Anitescu, C., Goswami, S., Nguyen-Thanh, V.~M., Guo, H., Hamdia, K., Zhuang, X., and Rabczuk, T. (2020).
\newblock An energy approach to the solution of partial differential equations in computational mechanics via machine learning: Concepts, implementation and applications.
\newblock {\em Computer Methods in Applied Mechanics and Engineering}, 362:112790.

\bibitem[Simard et~al., 1991]{LeCun91}
Simard, P.~Y., Victorri, B., LeCun, Y., and Denker, J.~S. (1991).
\newblock Tangent prop - {A} formalism for specifying selected invariances in an adaptive network.
\newblock In Moody, J.~E., Hanson, S.~J., and Lippmann, R., editors, {\em Advances in Neural Information Processing Systems 4, {[NIPS} Conference, Denver, Colorado, USA, December 2-5, 1991]}, pages 895--903. Morgan Kaufmann.

\bibitem[Srivastava et~al., 2014]{srivastava2014dropout}
Srivastava, N., Hinton, G., Krizhevsky, A., Sutskever, I., and Salakhutdinov, R. (2014).
\newblock Dropout: A simple way to prevent neural networks from overfitting.
\newblock {\em Journal of Machine Learning Research}, 15(56):1929--1958.

\bibitem[Taunk et~al., 2019]{nnalgo}
Taunk, K., De, S., Verma, S., and Swetapadma, A. (2019).
\newblock A brief review of nearest neighbor algorithm for learning and classification.
\newblock In {\em 2019 International Conference on Intelligent Computing and Control Systems (ICCS)}, pages 1255--1260.

\bibitem[Tibshirani, 1996]{tibshirani1996L1}
Tibshirani, R. (1996).
\newblock Regression shrinkage and selection via the lasso.
\newblock {\em Journal of the Royal Statistical Society: Series B (Methodological)}, 58(1):267--288.

\bibitem[Tikhonov, 1943]{TikhonovL2}
Tikhonov, A.~N. (1943).
\newblock On the stability of inverse problems.
\newblock {\em Proceedings of the USSR Academy of Sciences}, 39:195--198.

\bibitem[Tzen and Raginsky, 2019]{tzen2019neural}
Tzen, B. and Raginsky, M. (2019).
\newblock Neural stochastic differential equations: Deep latent gaussian models in the diffusion limit.
\newblock {\em arXiv preprint arXiv:1905.09883}.

\bibitem[Williams, 1995]{williams1995bayesian}
Williams, P.~M. (1995).
\newblock Bayesian regularization and pruning using a laplace prior.
\newblock {\em Neural computation}, 7(1):117--143.

\end{thebibliography}

%%%%%%%%%%%%%%%%%%%%%%%%%%%%%%%%%%%%%%%%%%%%%%%%%%%%%%%%%%%%%%%%%%%%%%%%%%%%%%%
%%%%%%%%%%%%%%%%%%%%%%%%%%%%%%%%%%%%%%%%%%%%%%%%%%%%%%%%%%%%%%%%%%%%%%%%%%%%%%%
% APPENDIX
%%%%%%%%%%%%%%%%%%%%%%%%%%%%%%%%%%%%%%%%%%%%%%%%%%%%%%%%%%%%%%%%%%%%%%%%%%%%%%%
%%%%%%%%%%%%%%%%%%%%%%%%%%%%%%%%%%%%%%%%%%%%%%%%%%%%%%%%%%%%%%%%%%%%%%%%%%%%%%%
\newpage
\appendix
\onecolumn

\begin{figure}
  \begin{center}
  \vskip -0.3in
  
  \centerline{
  \includegraphics[trim={0cm 1.2cm 2cm 2.4cm},clip, scale=0.27]{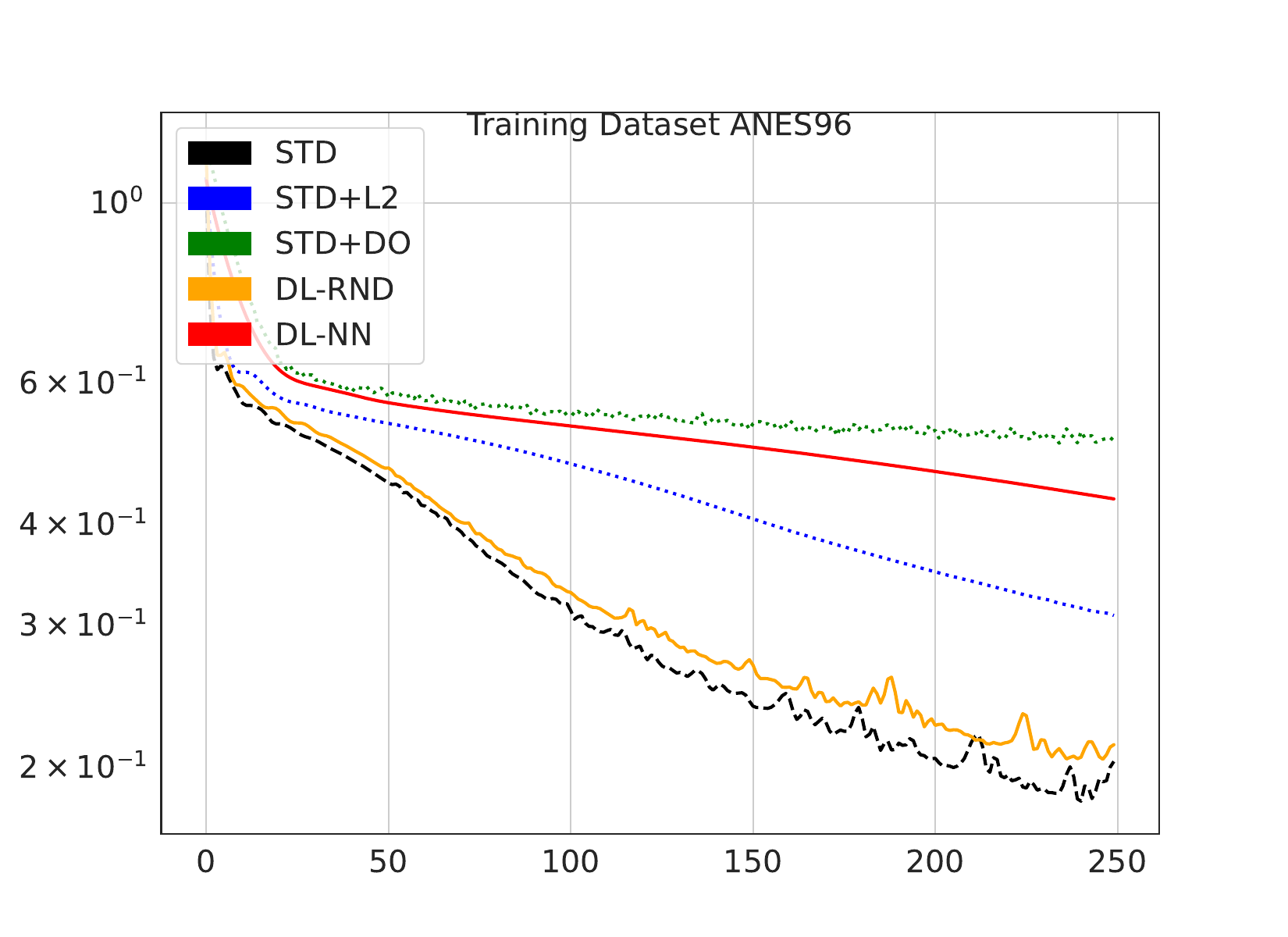}
  \includegraphics[trim={0cm 1.2cm 2cm 2.4cm},clip, scale=0.27]{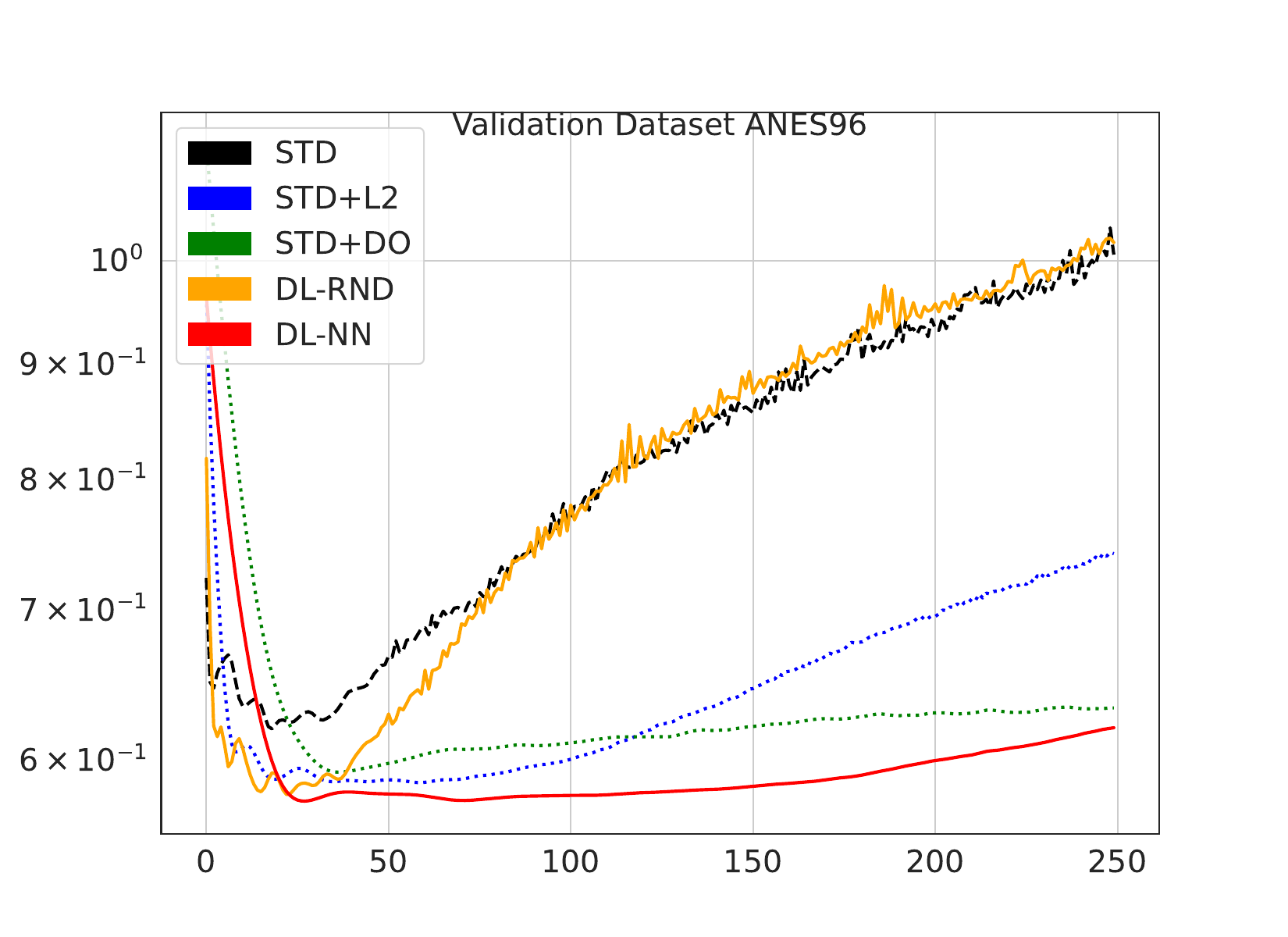}
  }
  
  \centerline{
  \includegraphics[trim={0cm 1.2cm 2cm 2.4cm},clip, scale=0.27]{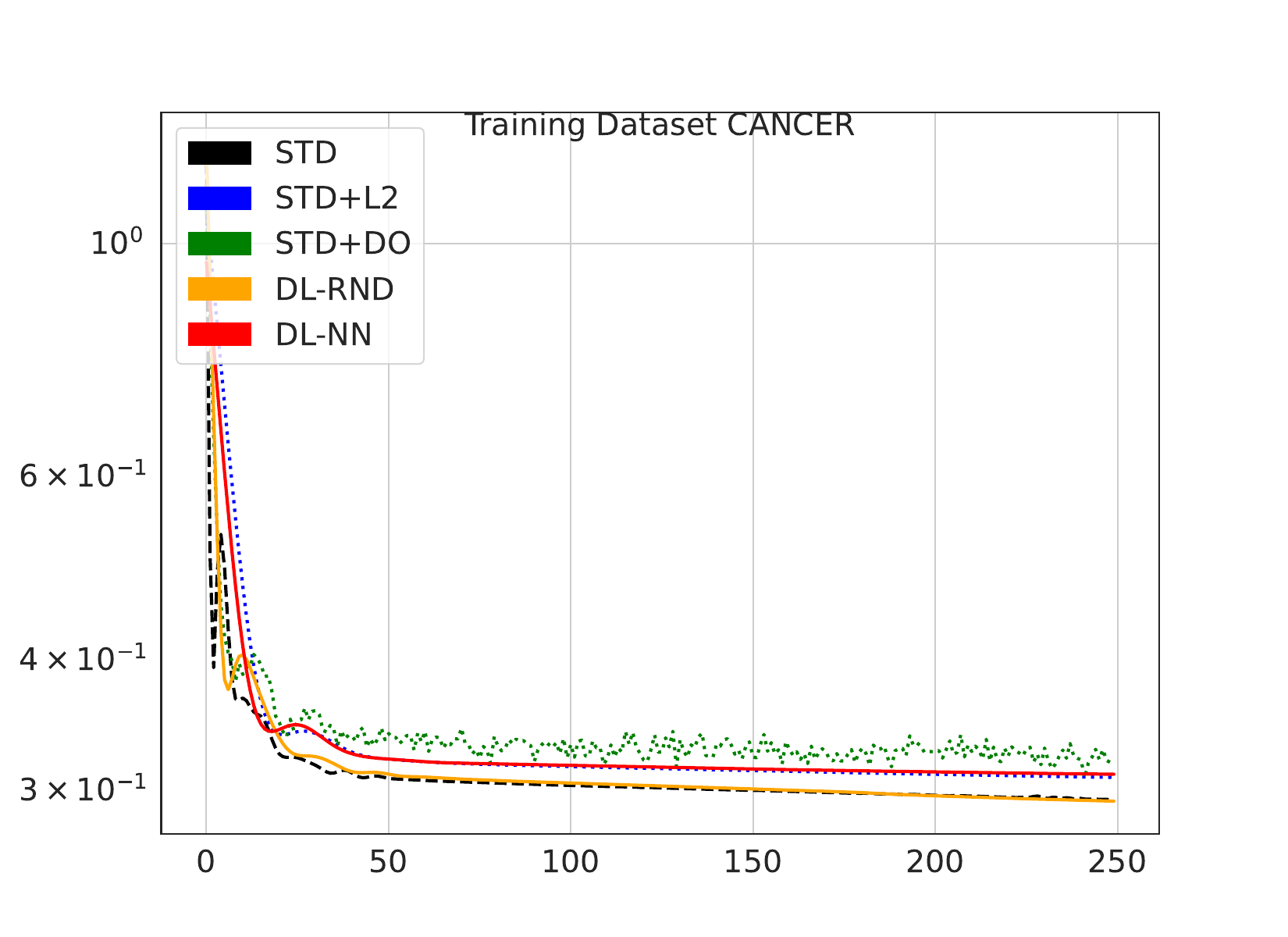}
  \includegraphics[trim={0cm 1.2cm 2cm 2.4cm},clip, scale=0.27]{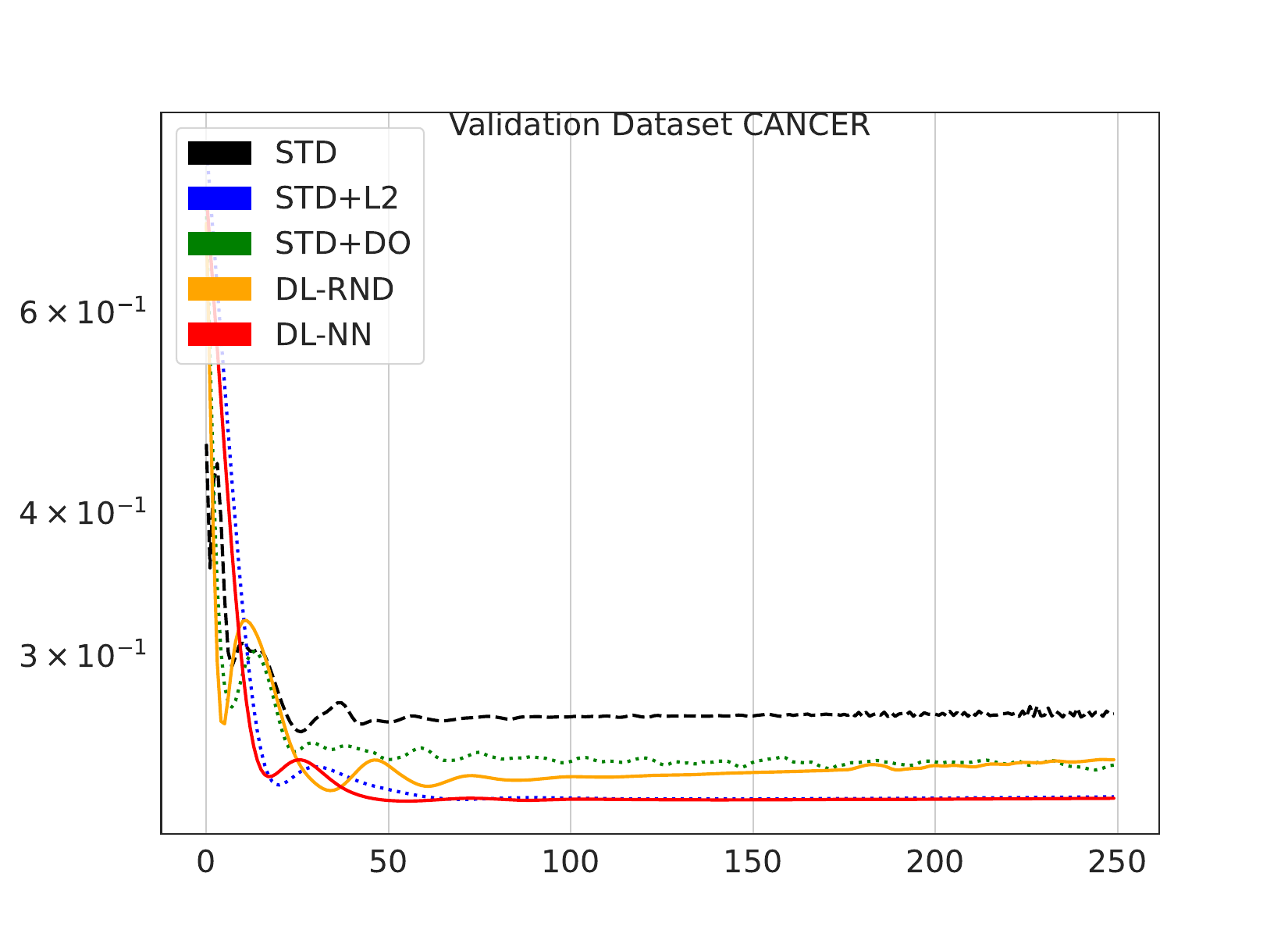}
  }
  
  \centerline{
  \includegraphics[trim={0cm 1.2cm 2cm 2.4cm},clip, scale=0.27]{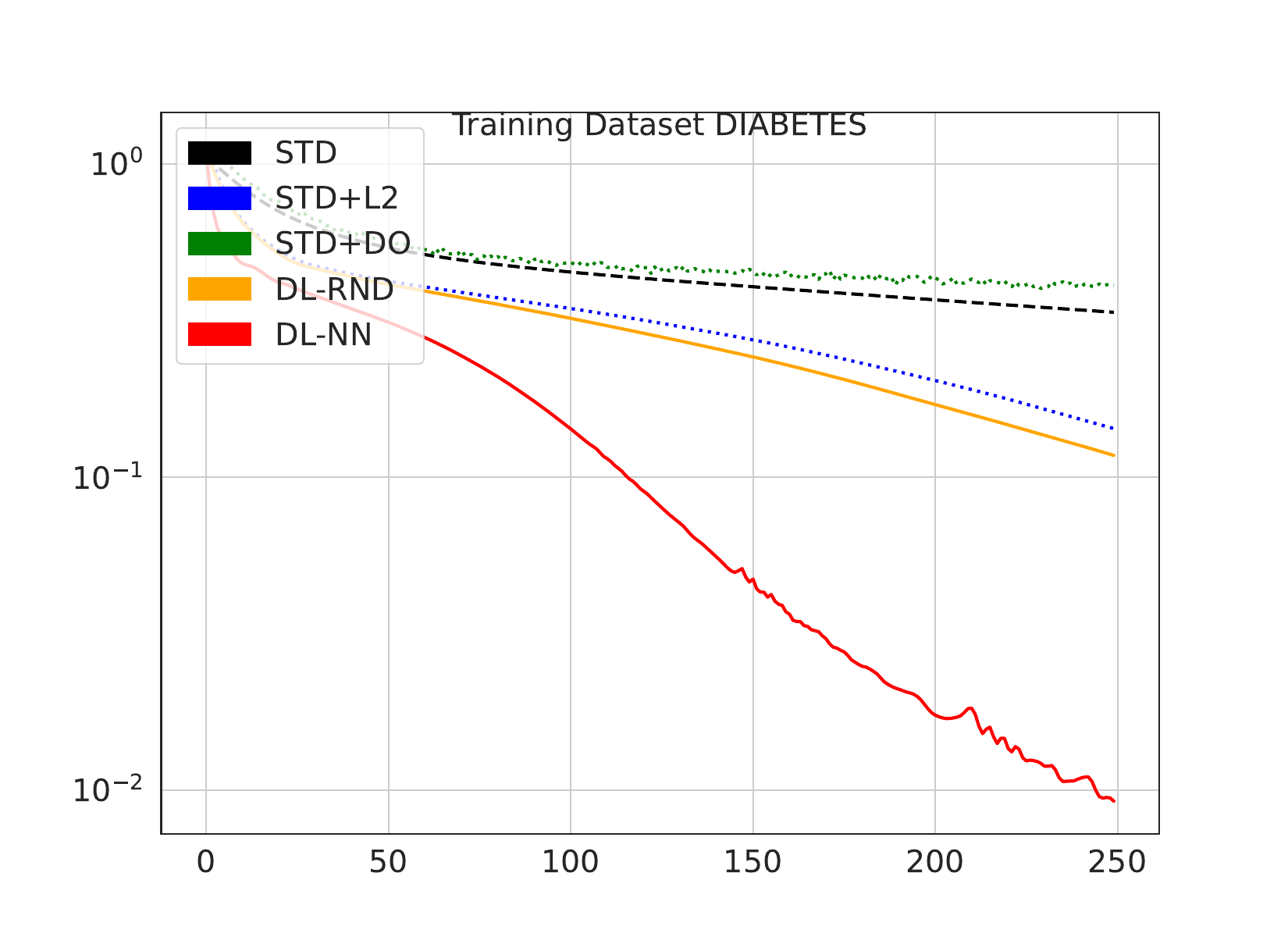}
  \includegraphics[trim={0cm 1.2cm 2cm 2.4cm},clip, scale=0.27]{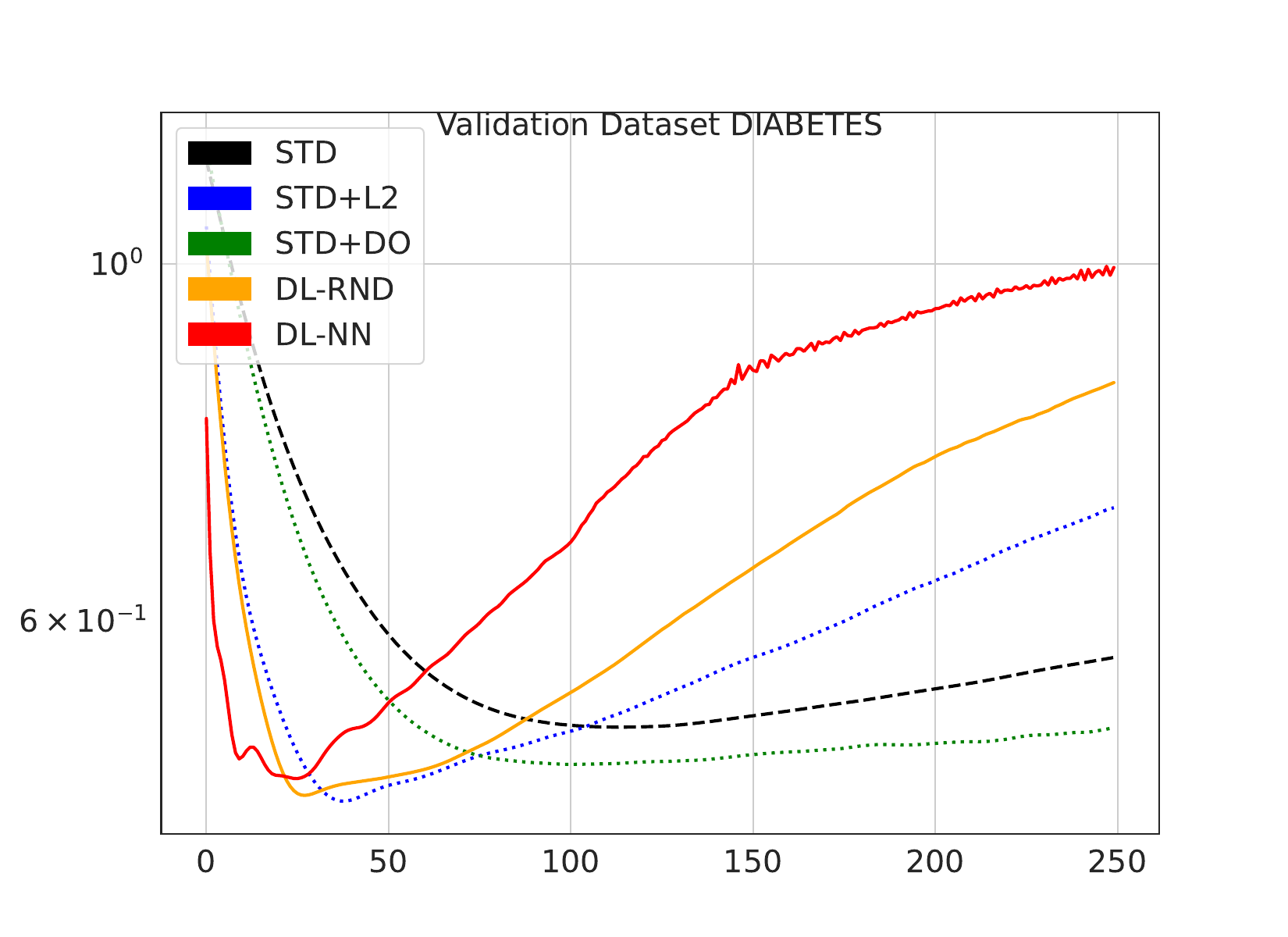}
  }

  \centerline{
  \includegraphics[trim={0cm 1.2cm 2cm 2.4cm},clip, scale=0.27]{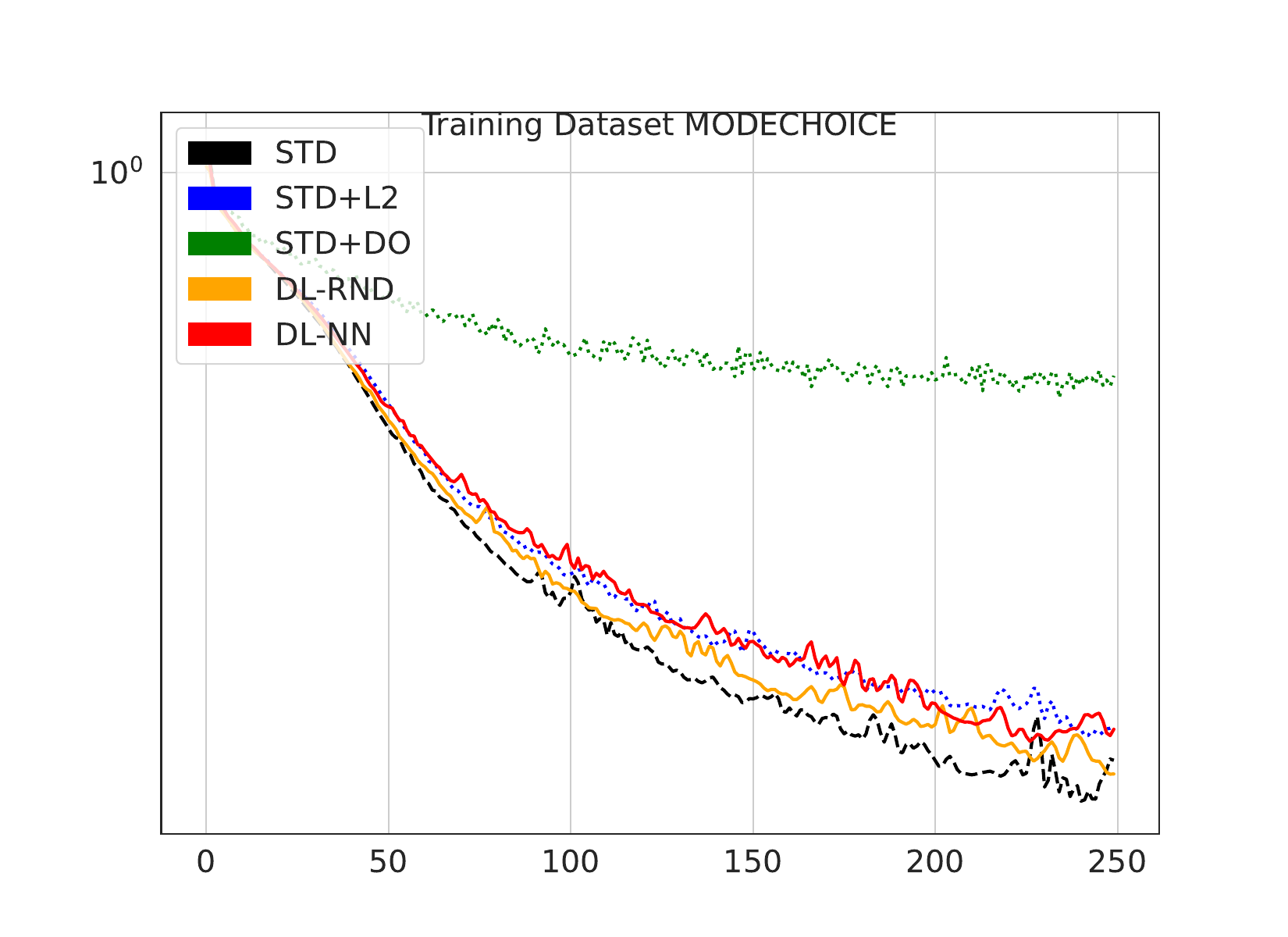}
  \includegraphics[trim={0cm 1.2cm 2cm 2.4cm},clip, scale=0.27]{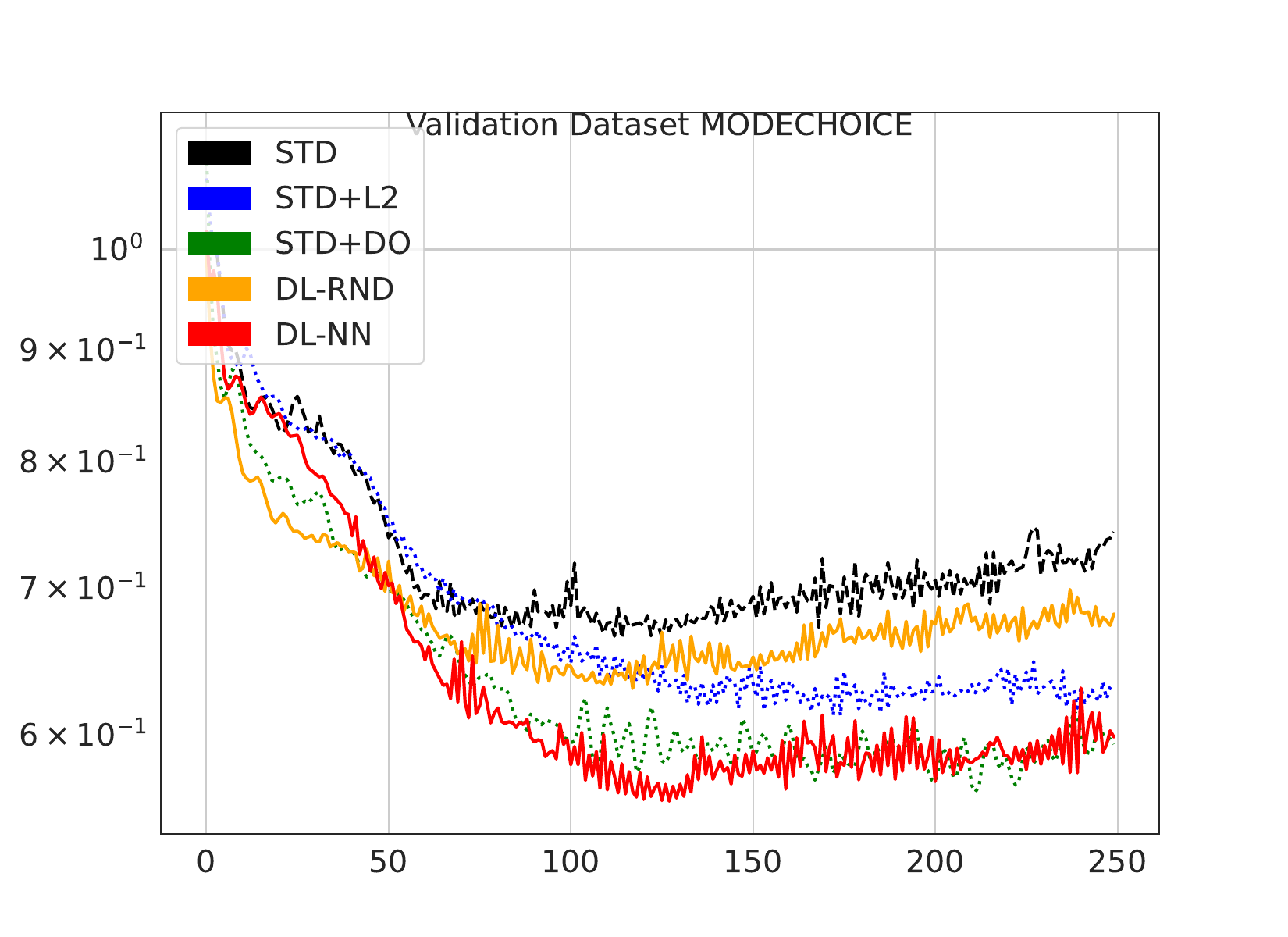}
  }
  
  \centerline{
  \includegraphics[trim={0cm 1.2cm 2cm 2.4cm},clip, scale=0.27]{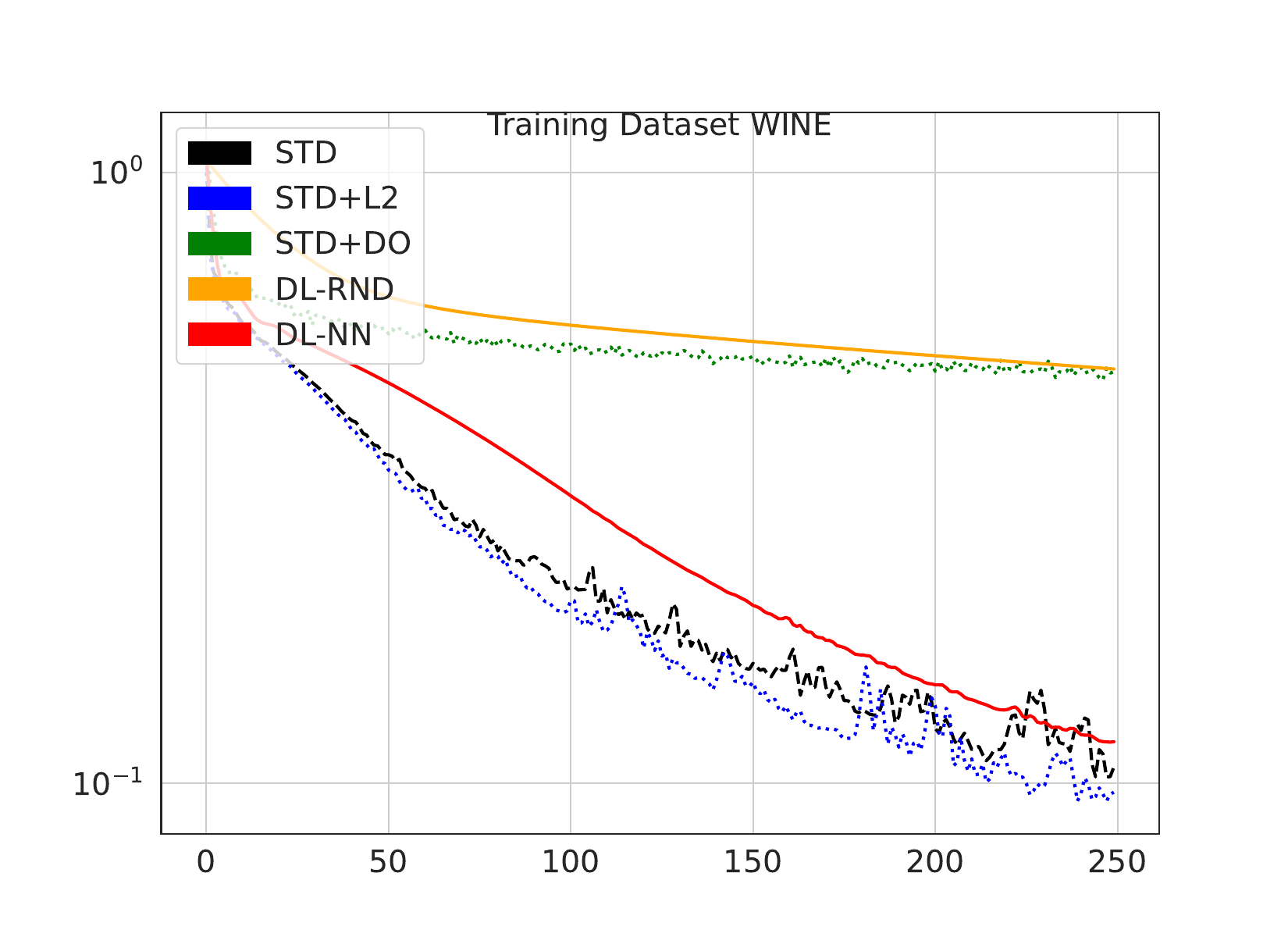}
  \includegraphics[trim={0cm 1.2cm 2cm 2.4cm},clip, scale=0.27]{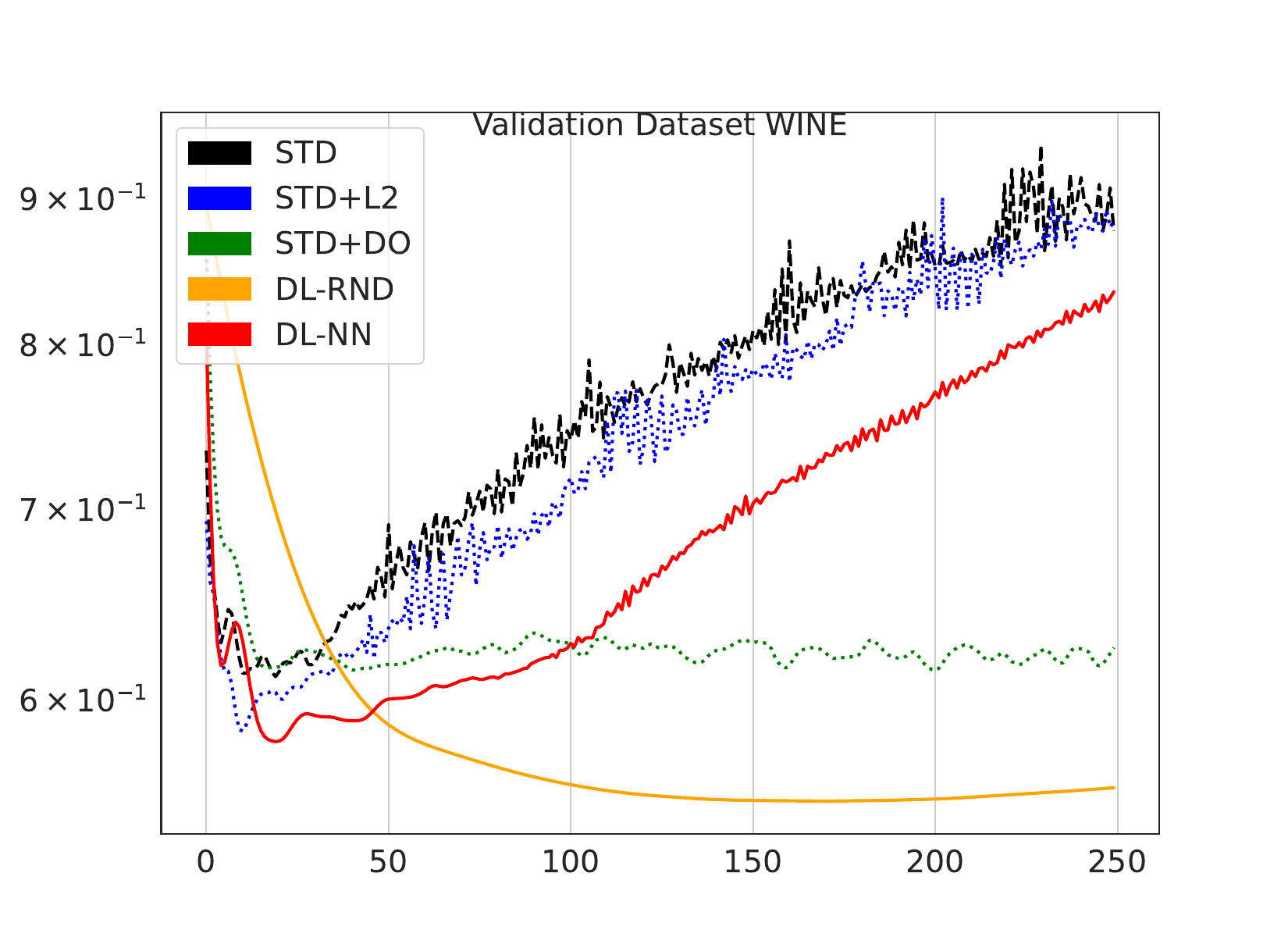}
  }
  
  \vskip -0.1in
  \caption{Learning curves of the 5-folds cross validation average fro Real Dataset group. 
  Training set - left - and Validation set - right.
  Epochs on the x-axis and MSE on the y-axes.
  The curves show: $STD$, $STD+L_2$, $STD+DO$, $DL_{RND}$ and $DL_{NN}$, each with the best parameters.
  Selection criterion is the parameter combination leading to lowest $MSE_{val}$. 
  Real datasets: ANES96, CANCER, DIABETES, MODECHOICE and WINE.
  }
  \label{fig:learn-real}
  \end{center}
\end{figure}

%%%%%%%%%%%%%%%%%%%%%%%%%%%

\begin{figure}
  \begin{center}
  
  \vskip -0.3in
  
  \centerline{
  \includegraphics[trim={0cm 1.2cm 2cm 2.4cm},clip, scale=0.27]{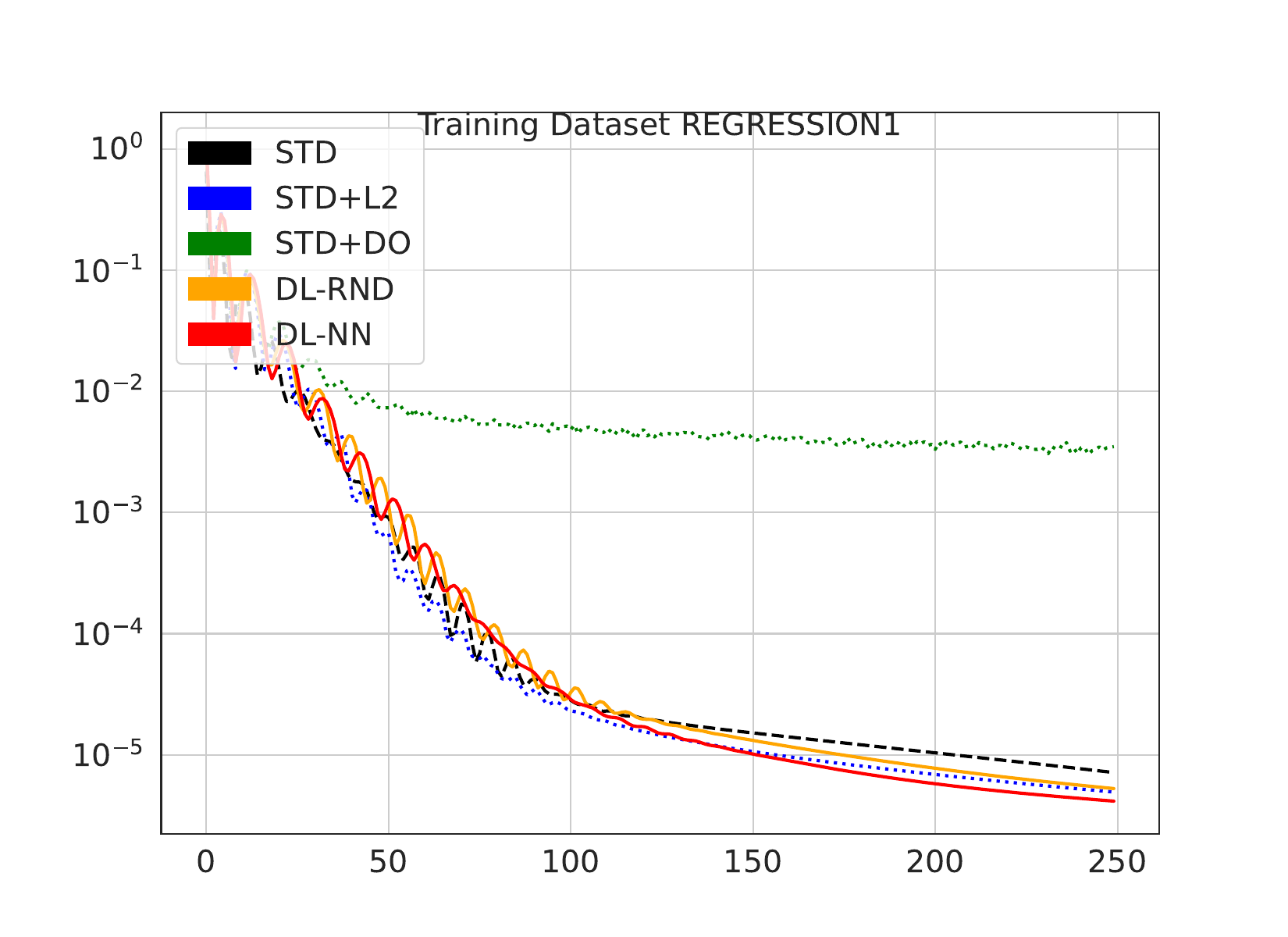}
  \includegraphics[trim={0cm 1.2cm 2cm 2.4cm},clip, scale=0.27]{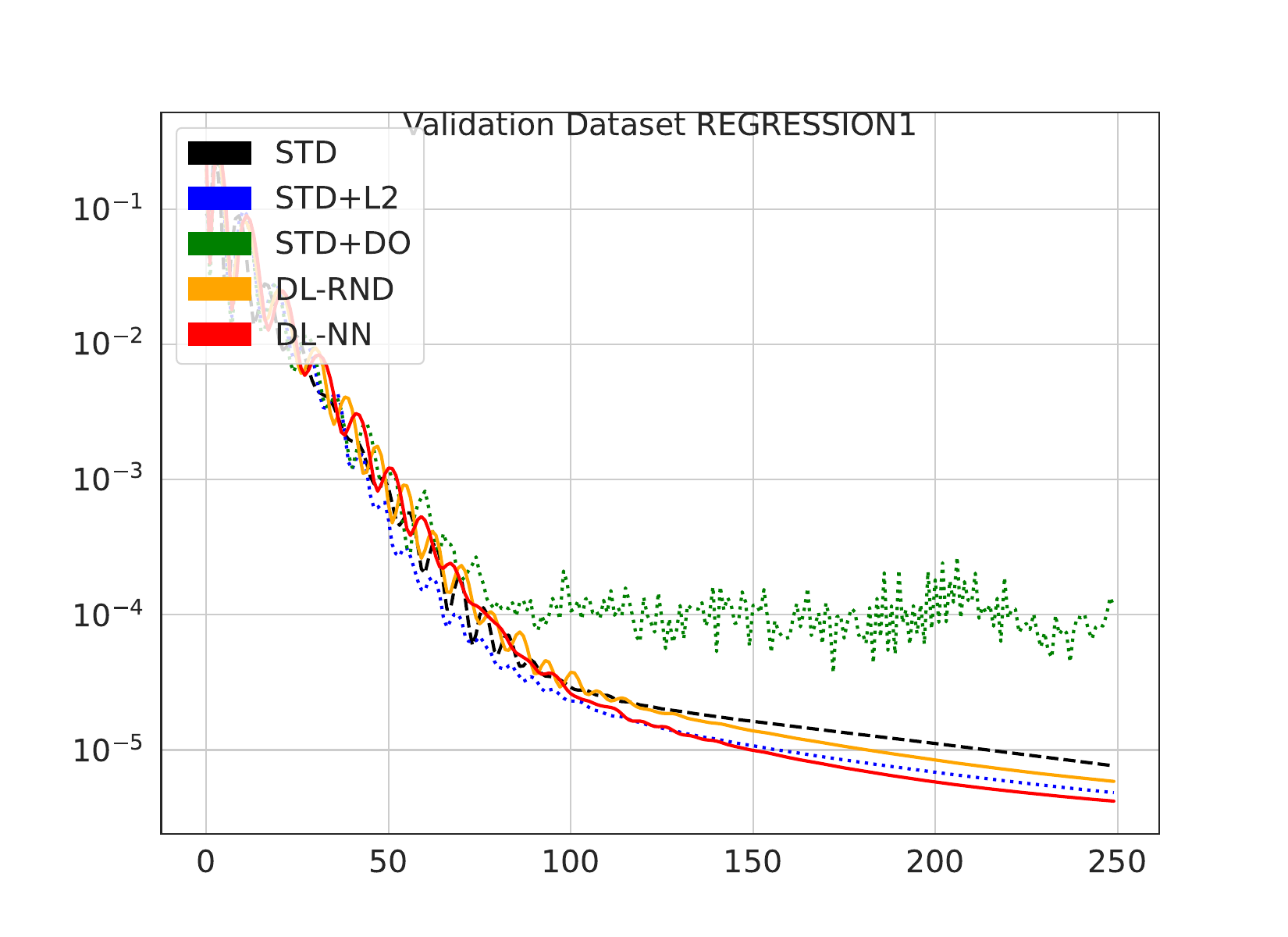}
  }
  
  \centerline{
  \includegraphics[trim={0cm 1.2cm 2cm 2.4cm},clip, scale=0.27]{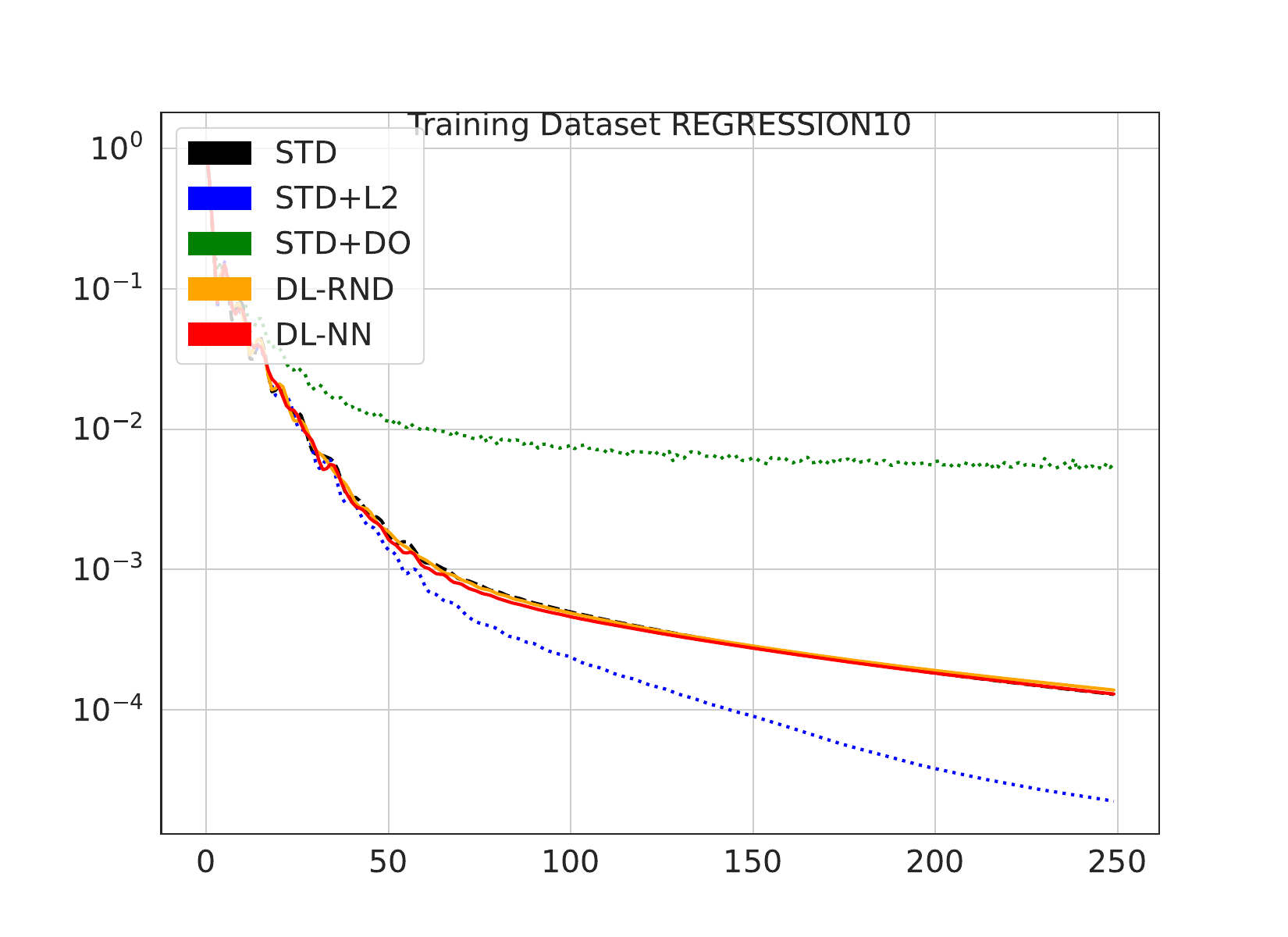}
  \includegraphics[trim={0cm 1.2cm 2cm 2.4cm},clip, scale=0.27]{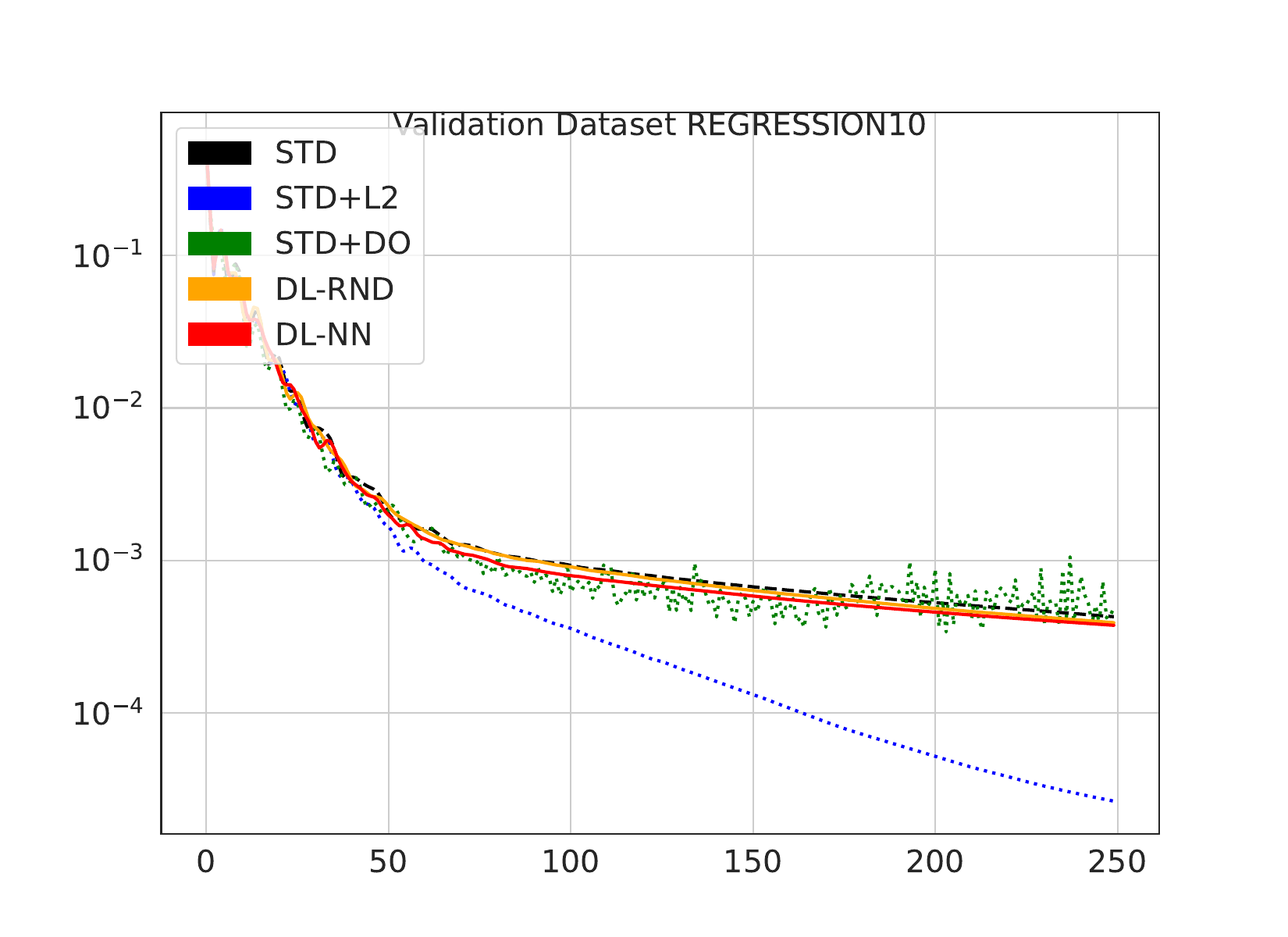}
  }
  
  \centerline{
  \includegraphics[trim={0cm 1.2cm 2cm 2.4cm},clip, scale=0.27]{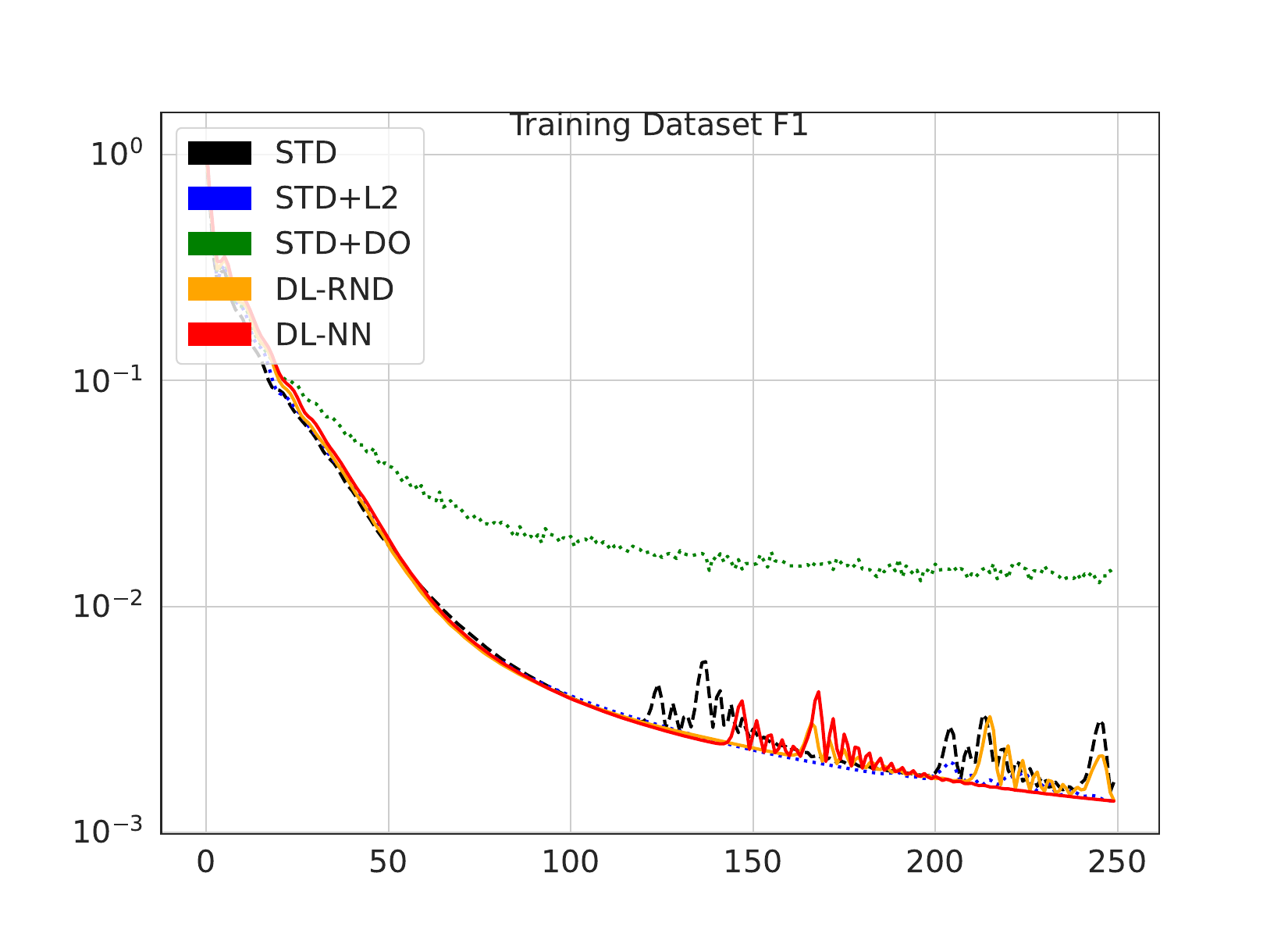}
  \includegraphics[trim={0cm 1.2cm 2cm 2.4cm},clip, scale=0.27]{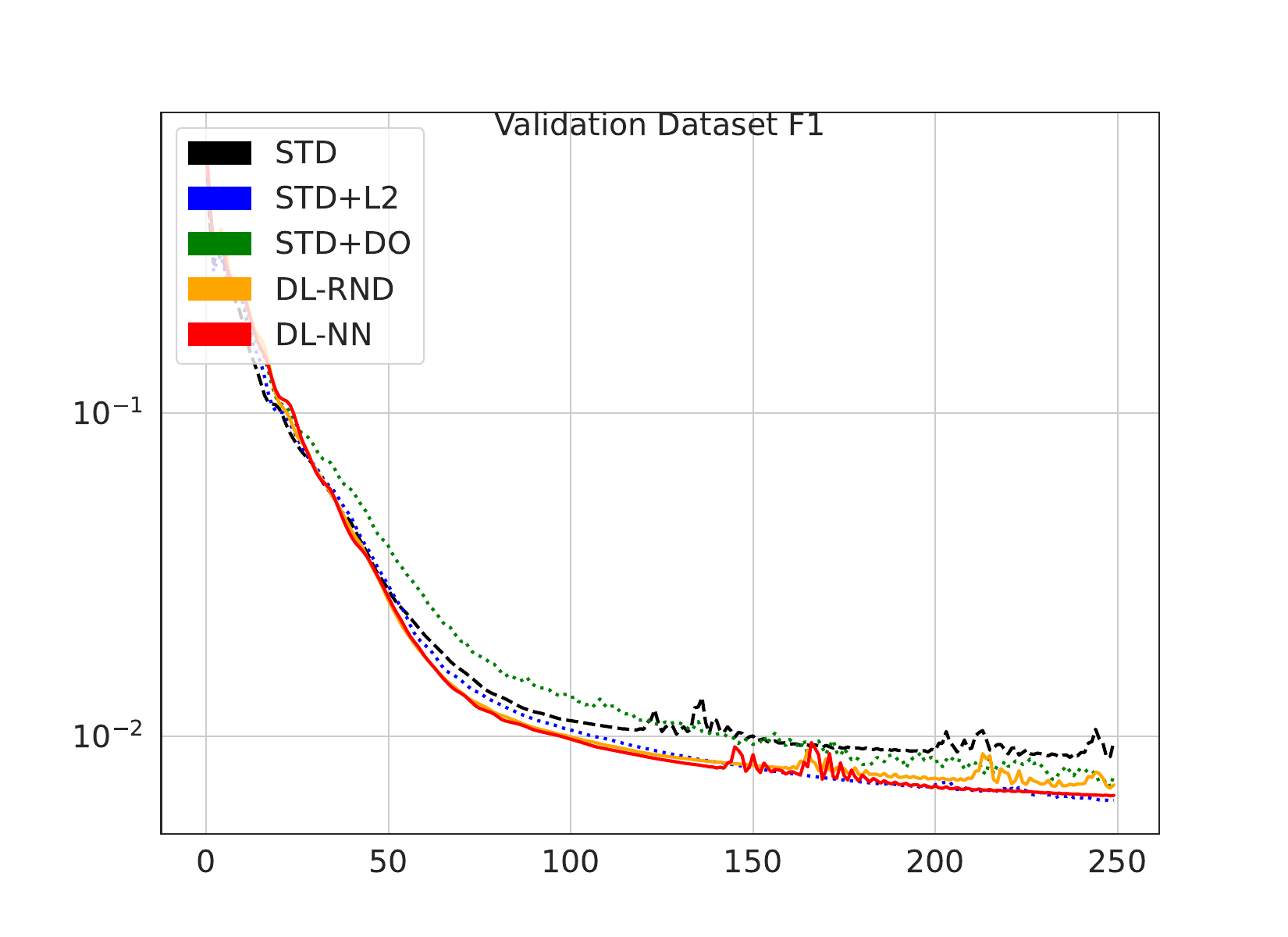}
  }
  
  \centerline{
  \includegraphics[trim={0cm 1.2cm 2cm 2.4cm},clip, scale=0.27]{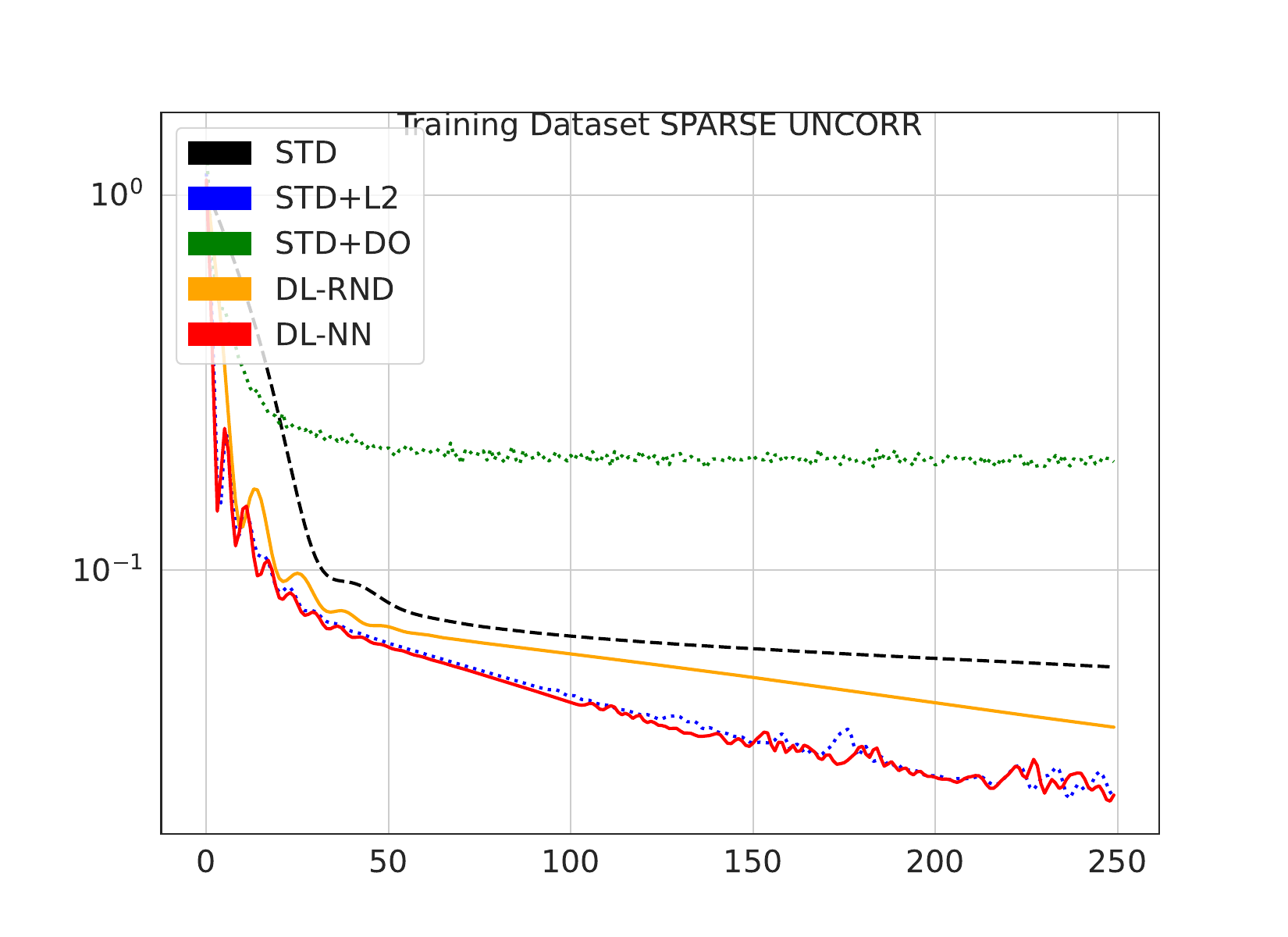}
  \includegraphics[trim={0cm 1.2cm 2cm 2.4cm},clip, scale=0.27]{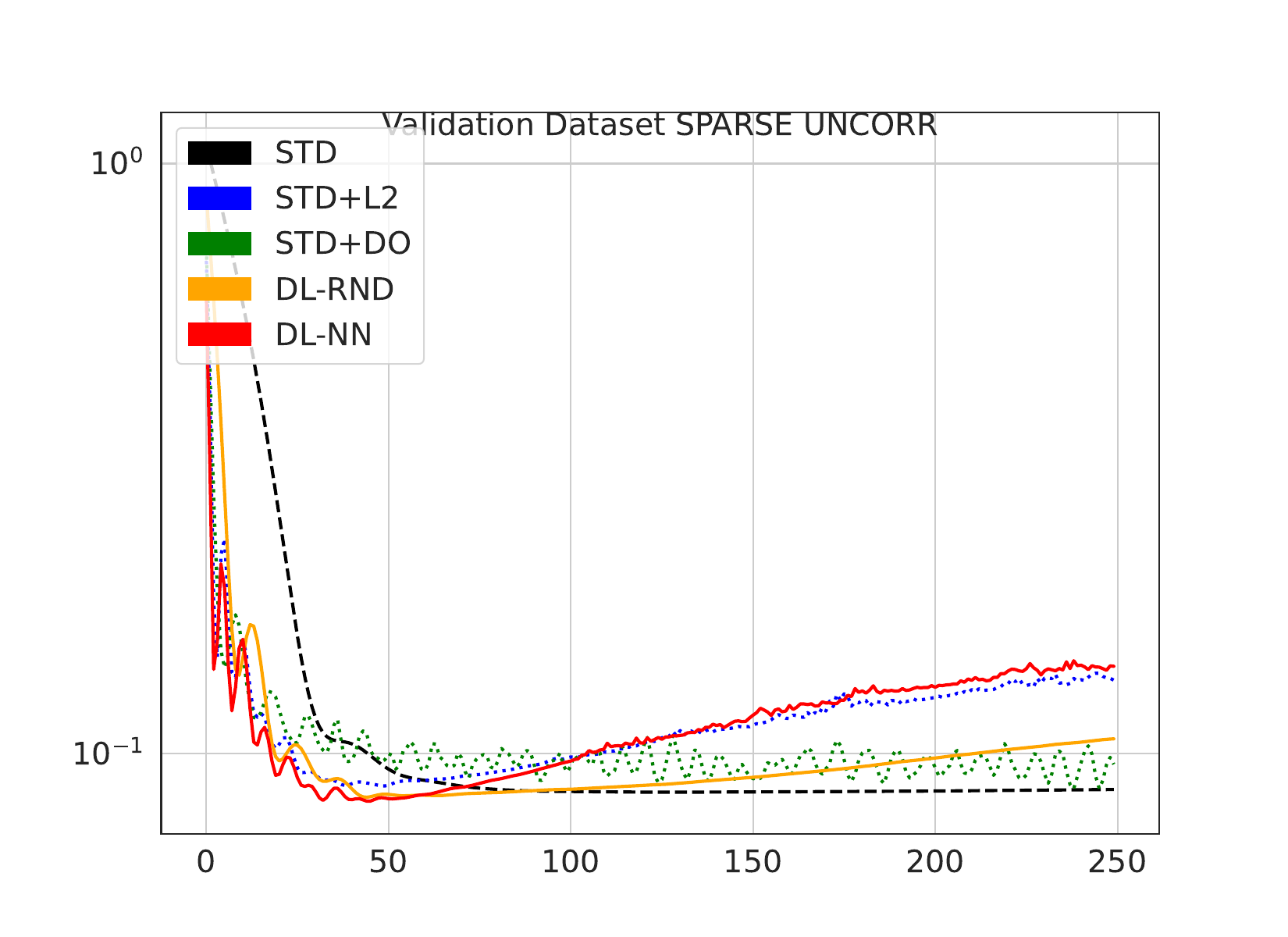}
  }
  
  \centerline{
  \includegraphics[trim={0cm 1.2cm 2cm 2.4cm},clip, scale=0.27]{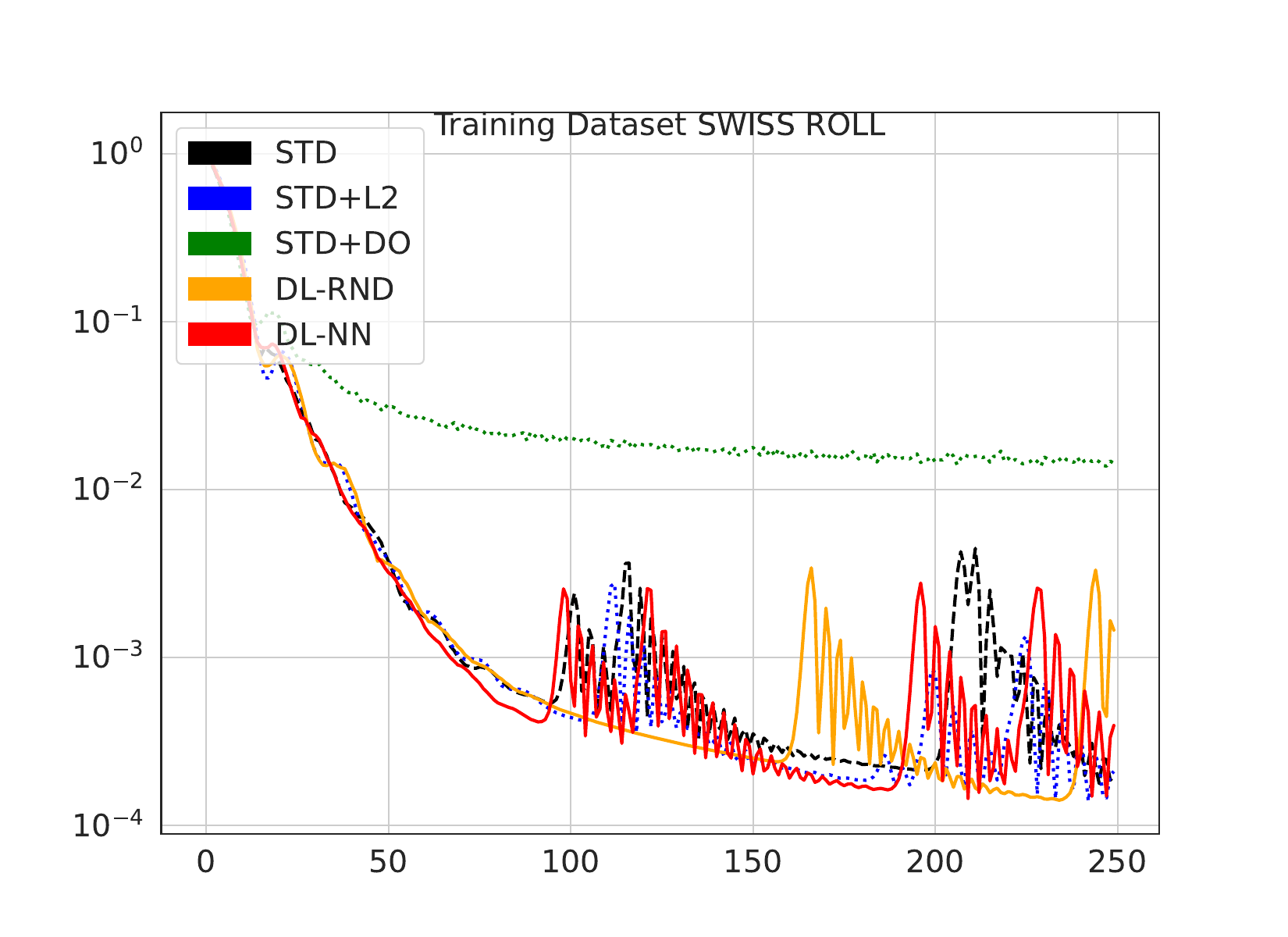}
  \includegraphics[trim={0cm 1.2cm 2cm 2.4cm},clip, scale=0.27]{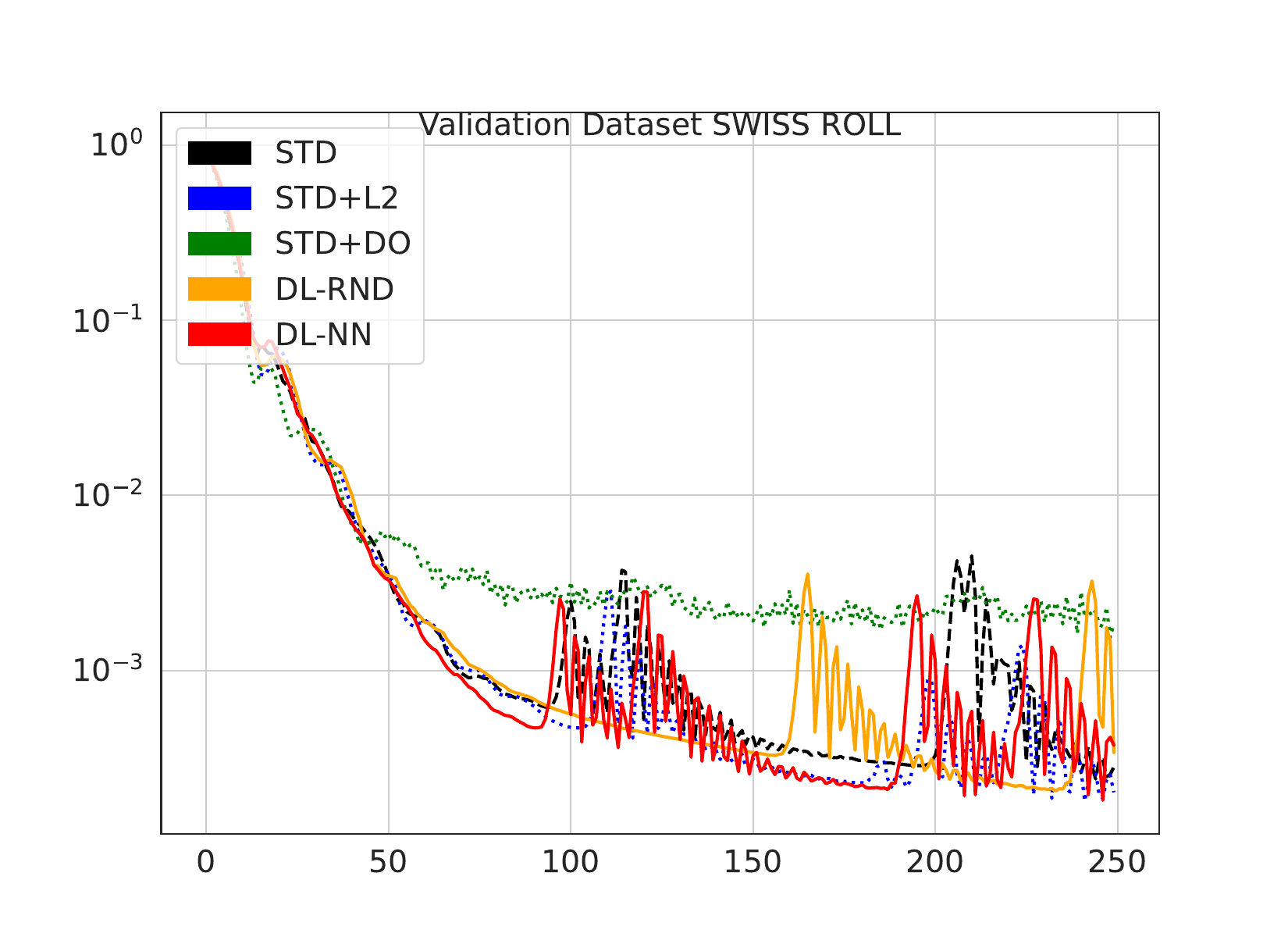}
  }
  
  \vskip -0.1in
  \caption{Learning curves of the 5-folds cross validation average fro Synthetic Dataset group. 
  Training set - left - and Validation set - right.
  Epochs on the x-axis and MSE on the y-axes.
  The curves show: $STD$, $STD+L_2$, $STD+DO$, $DL_{RND}$ and $DL_{NN}$, each with the best parameters.
  Selection criterion is the parameter combination leading to lowest $MSE_{val}$.
  Synthetic datasets: F1, REGRESSION1 and 10, SPARSE UNCORR and SWISS ROLL.}
  \label{fig:learn-syn}
  \end{center}
\end{figure}

%%%%%%%%%%%%%%%%%%%%%%%%%%%%%%%%%%%%%%%%%%%%%%%%%%%%%%%%%%%%%%%%%%%%%%%%%%%%%%%

\begin{table}[tb]
  \begin{center}
  \begin{small}
  \begin{sc}
  
  \begin{tabular}{lllllll}
  \toprule
  Data & Method & $L2$ & tuples & $\theta_{D}$ & $\lambda$ & Dropout($p$) \\

  \midrule
\multirow{5}{*}{anes96}  & STD  & && & 0.03 &  \\
 & STD+$L_2$& $1 \times 10^{-3}$ & & & 0.01 &  \\
 & STD+DO & &&& 0.003 & 0.1  \\
 & $DL_{RND}$ & & 1  & $1	\times 10^{-7}$ & 0.03 & \\
 & $DL_{NN}$ & & 3  & $1 	\times 10^{-6}$ & 0.003 & \\ \cline{2-7}
\multirow{5}{*}{cancer}  & STD  & & && 0.03 & \\
 & STD+$L_2$& $1 \times 10^{-5}$ & && 0.003 & \\
 & STD+DO & & & & 0.01 & 0.1  \\
 & $DL_{RND}$ & & 1  & $1	\times 10^{-7}$ & 0.01 & \\
 & $DL_{NN}$ & & 1  & $1	\times 10^{-7}$ & 0.003 & \\\cline{2-7}
\multirow{5}{*}{diabetes} & STD  & & & & $1 \times 10^{-3}$ & \\
 & STD+$L_2$ & $1 	\times 10^{-6}$ & & & 0.003 & \\
 & STD+DO & & & & $1 \times 10^{-3}$ & 0.2  \\
 & $DL_{RND}$ & & 1  & $1 \times 10^{-4}$ & 0.003 & \\
 & $DL_{NN}$ & & 3  & $1 \times 10^{-3}$ & 0.01 & \\\cline{2-7}
\multirow{5}{*}{modechoice} & STD  & & & & 0.03 & \\
 & STD+$L_2$ & $1 \times 10^{-4}$ & & & 0.03 & \\
 & STD+DO & & & & 0.03 & 0.2  \\
 & $DL_{RND}$ & & 3  & $1	\times 10^{-7}$ & 0.03 & \\
 & $DL_{NN}$ & & 3  & $1 \times 10^{-3}$ & 0.03 & \\\cline{2-7}
\multirow{5}{*}{wine} & STD  & & & & 0.03 & \\
 & STD+$L_2$ & $1 	\times 10^{-6}$ & & & 0.03 & \\
 & STD+DO & & & & 0.01 & 0.4  \\
 & $DL_{RND}$ & & 3  & $1 	\times 10^{-6}$ & $1 \times 10^{-3}$ & \\
 & $DL_{NN}$ & & 1  & $1 \times 10^{-5}$ & 0.01 &  \\
 \hline
\multirow{5}{*}{f1} & STD  & & & & 0.03 & \\
 & STD+$L_2$ & $1 \times 10^{-4}$ & & & 0.03 & \\
 & STD+DO & & & & 0.03 & 0.05 \\
 & $DL_{RND}$ & & 1  & $1 \times 10^{-3}$ & 0.03 & \\
 & $DL_{NN}$ & & 3  & $1 \times 10^{-5}$ & 0.03 & \\\cline{2-7}
\multirow{5}{*}{regression1}  & STD  & & & & 0.03 & \\
 & STD+$L_2$ & $1 \times 10^{-5}$ & & & 0.03 & \\
 & STD+DO & & & & 0.03 & 0.05 \\
 & $DL_{RND}$ & & 1  & $1 \times 10^{-3}$ & 0.03 & \\
 & $DL_{NN}$ & & 3  & $1 \times 10^{-3}$ & 0.03 & \\\cline{2-7}
\multirow{5}{*}{regression10} & STD  & & & & 0.03 & \\
 & STD+$L_2$ & $1 \times 10^{-3}$ & & & 0.03 & \\
 & STD+DO & & & & 0.03 & 0.1  \\
 & $DL_{RND}$ & & 3  & $1 \times 10^{-5}$ & 0.03 & \\
 & $DL_{NN}$ & & 1  & $1 	\times 10^{-6}$ & 0.03 & \\\cline{2-7}
\multirow{5}{*}{sparse uncorr} & STD  & & & & 0.003 & \\
 & STD+$L_2$ & $1	\times 10^{-7}$ & & & 0.03 & \\
 & STD+DO & & & & 0.03 & 0.8  \\
 & $DL_{RND}$ & & 1  & $1 \times 10^{-3}$ & 0.01 & \\
 & $DL_{NN}$ & & 3  & $1 \times 10^{-5}$ & 0.03 & \\\cline{2-7}
\multirow{5}{*}{swiss roll} & STD  & & & & 0.03 & \\
 & STD+$L_2$ & $1 \times 10^{-4}$ & & & 0.03 & \\
 & STD+DO & & & & 0.03 & 0.05 \\
 & $DL_{RND}$ & & 3  & $1 \times 10^{-5}$ & 0.03 & \\
 & $DL_{NN}$ & & 1  & $1 \times 10^{-4}$ & 0.03 & \\

 \bottomrule
  \end{tabular}
  \end{sc}
  \end{small}
  \end{center}

  \caption{Parameters leading to best results for models $STD$, $STD+L_2$, $STD+DO$, $DL_{RND}$, and $DL_{NN}$ per each dataset.}
  \label{tab:bestparam-table}
\end{table}

%%%%%%%%%%%%%%%%%%%%%%%%%%%%%%%%%%%%%%%%%%%%%%%%%%%%%%%%%%%%%%%%%%%%%%%%%%%%%%%

\begin{figure}
  \begin{center}
  
  \centerline{
  \includegraphics[trim={2cm 0cm 3cm 2cm},clip, scale=0.25]{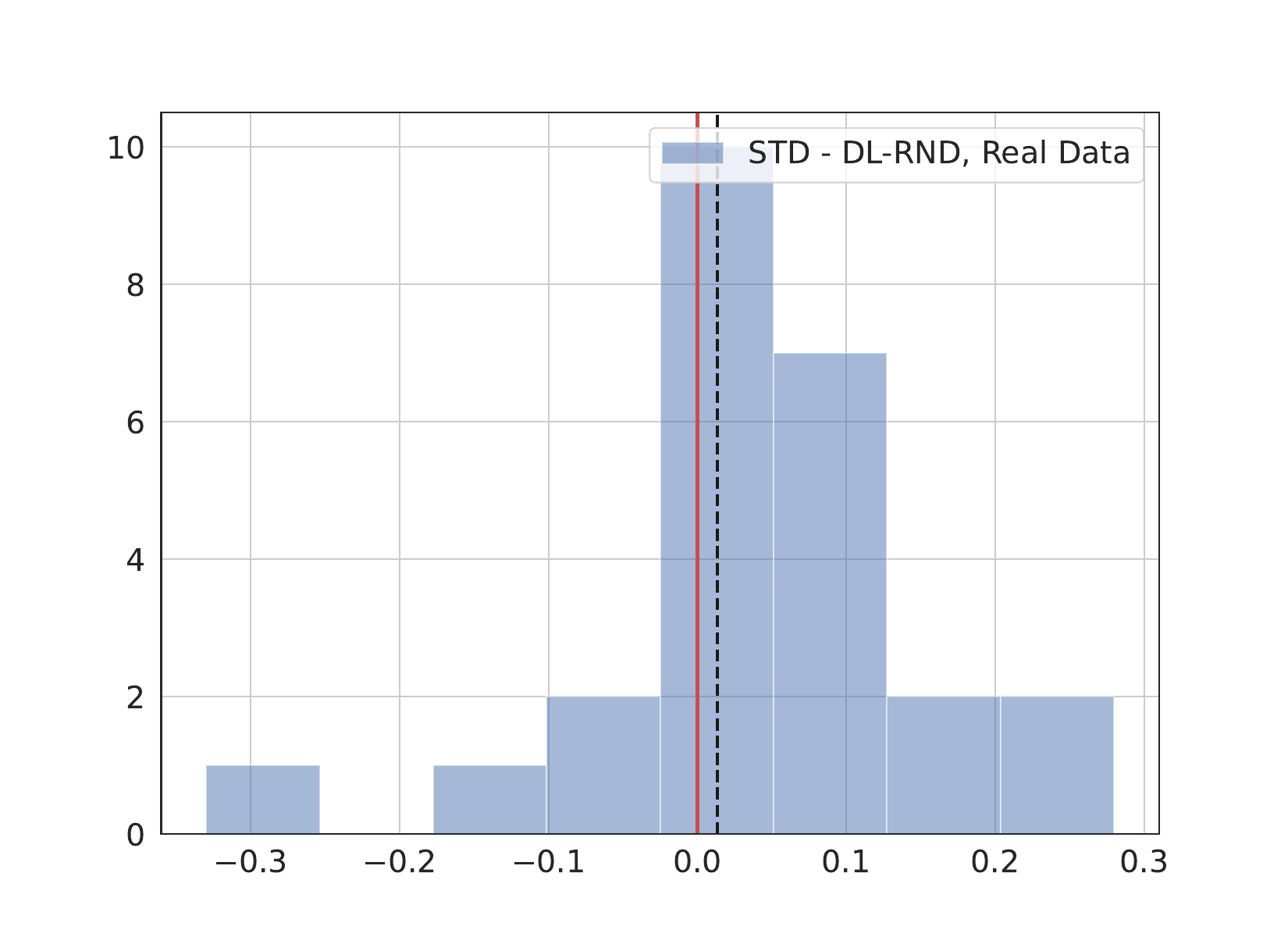}
  \includegraphics[trim={2cm 0cm 3cm 2cm},clip, scale=0.25]{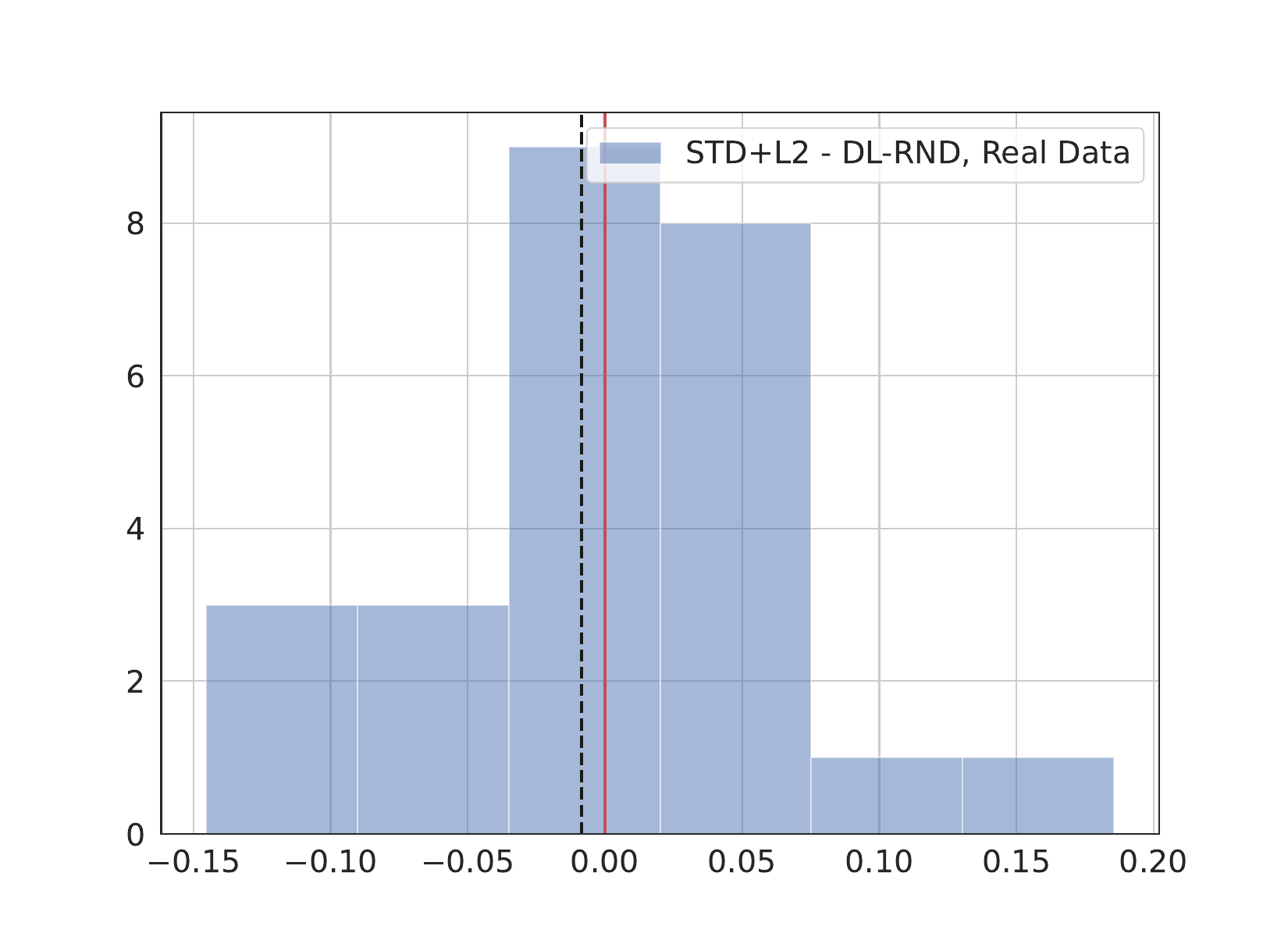}
  \includegraphics[trim={2cm 0cm 3cm 2cm},clip, scale=0.25]{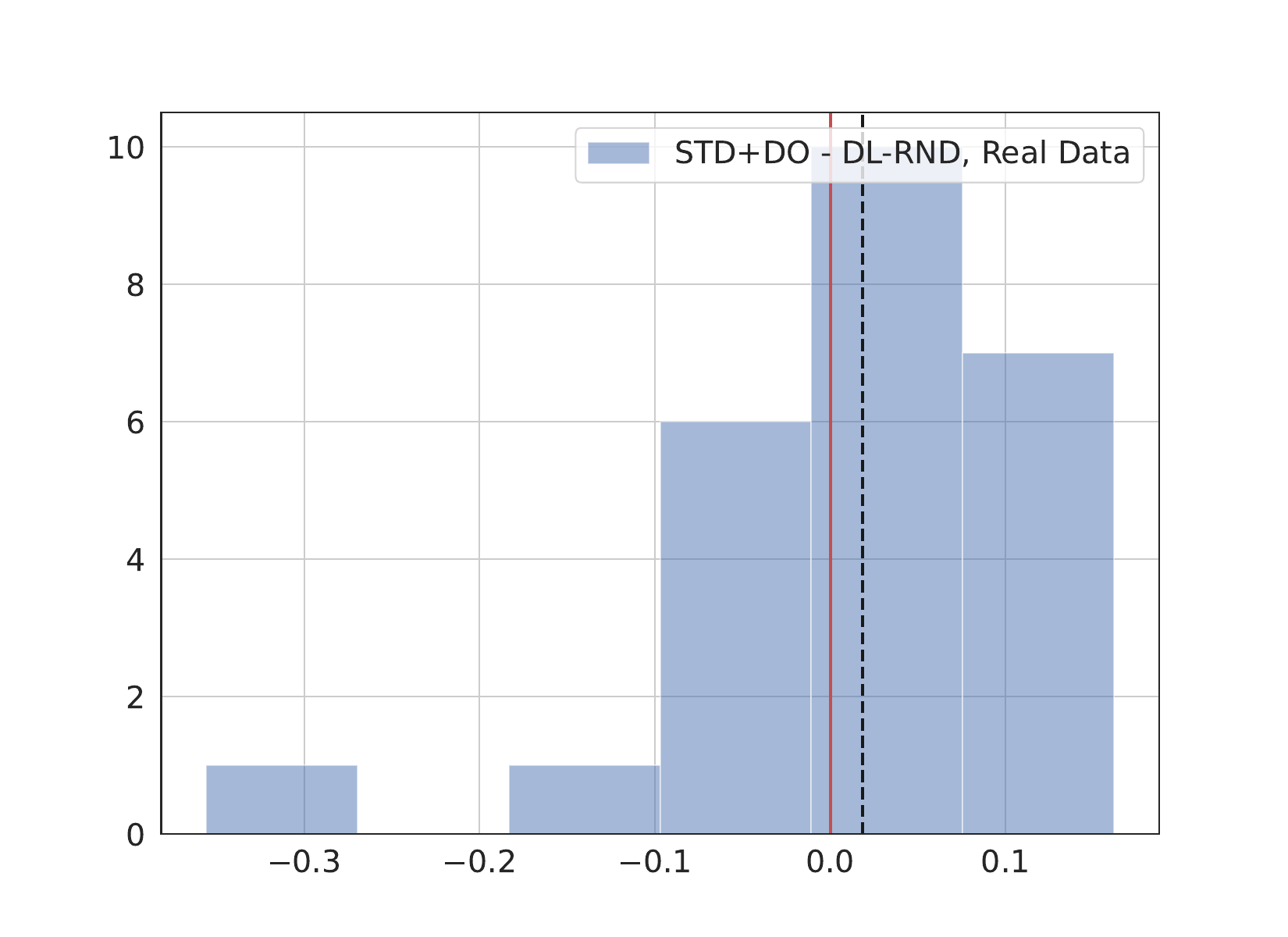}
  }
  \centerline{
  \includegraphics[trim={2cm 0cm 3cm 2cm},clip, scale=0.25]{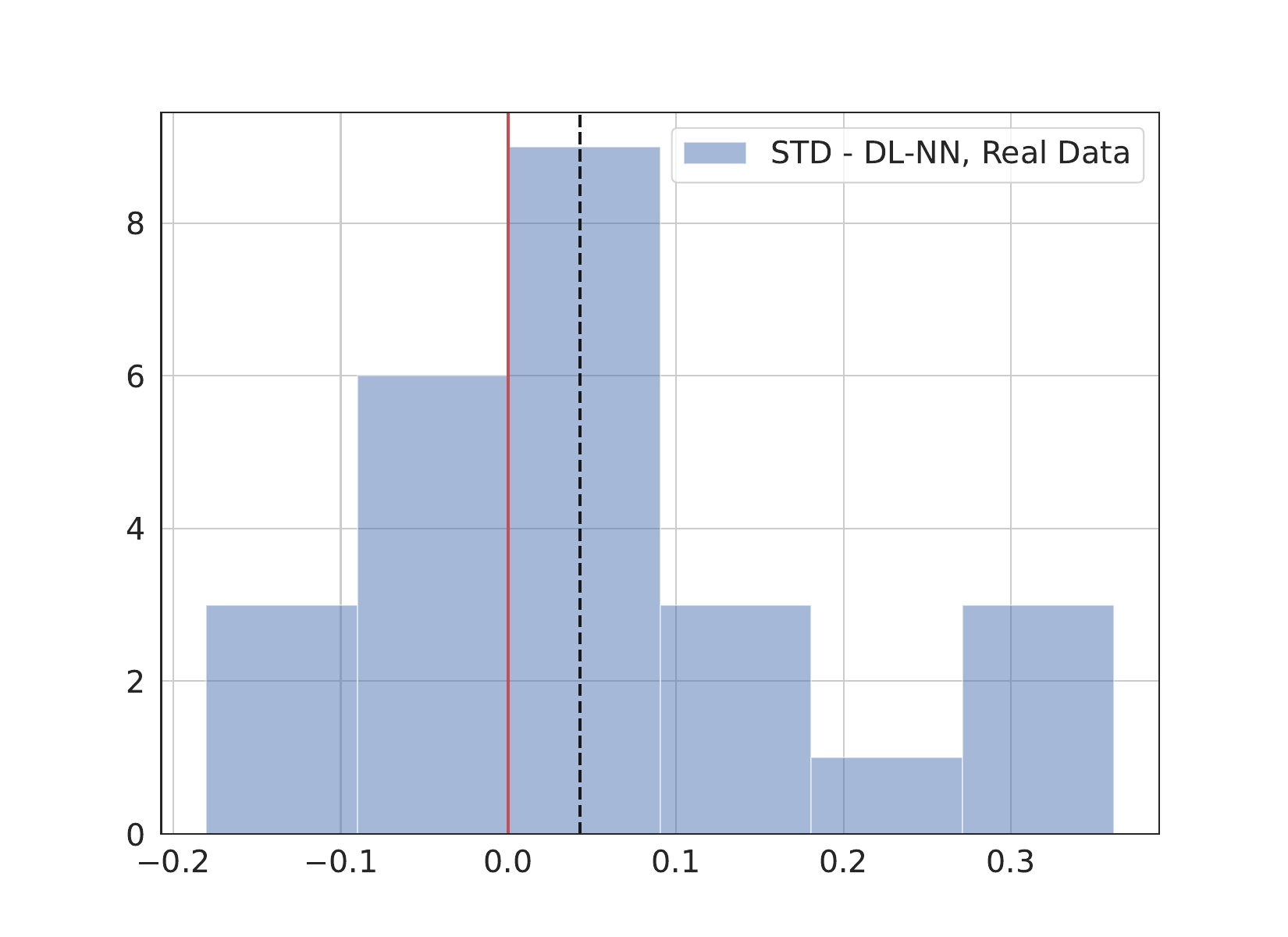}
  \includegraphics[trim={2cm 0cm 3cm 2cm},clip, scale=0.25]{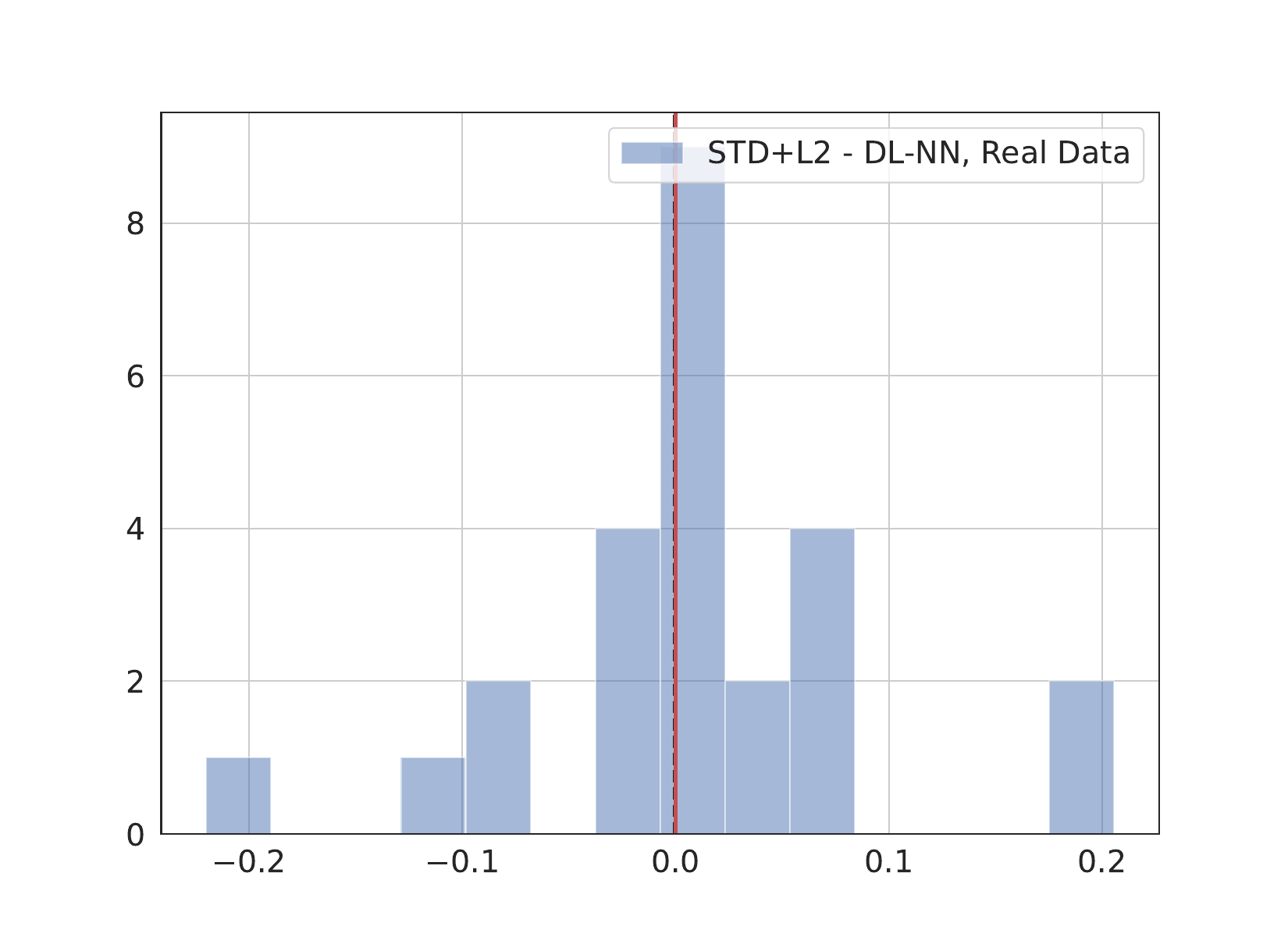}
  \includegraphics[trim={2cm 0cm 3cm 2cm},clip, scale=0.25]{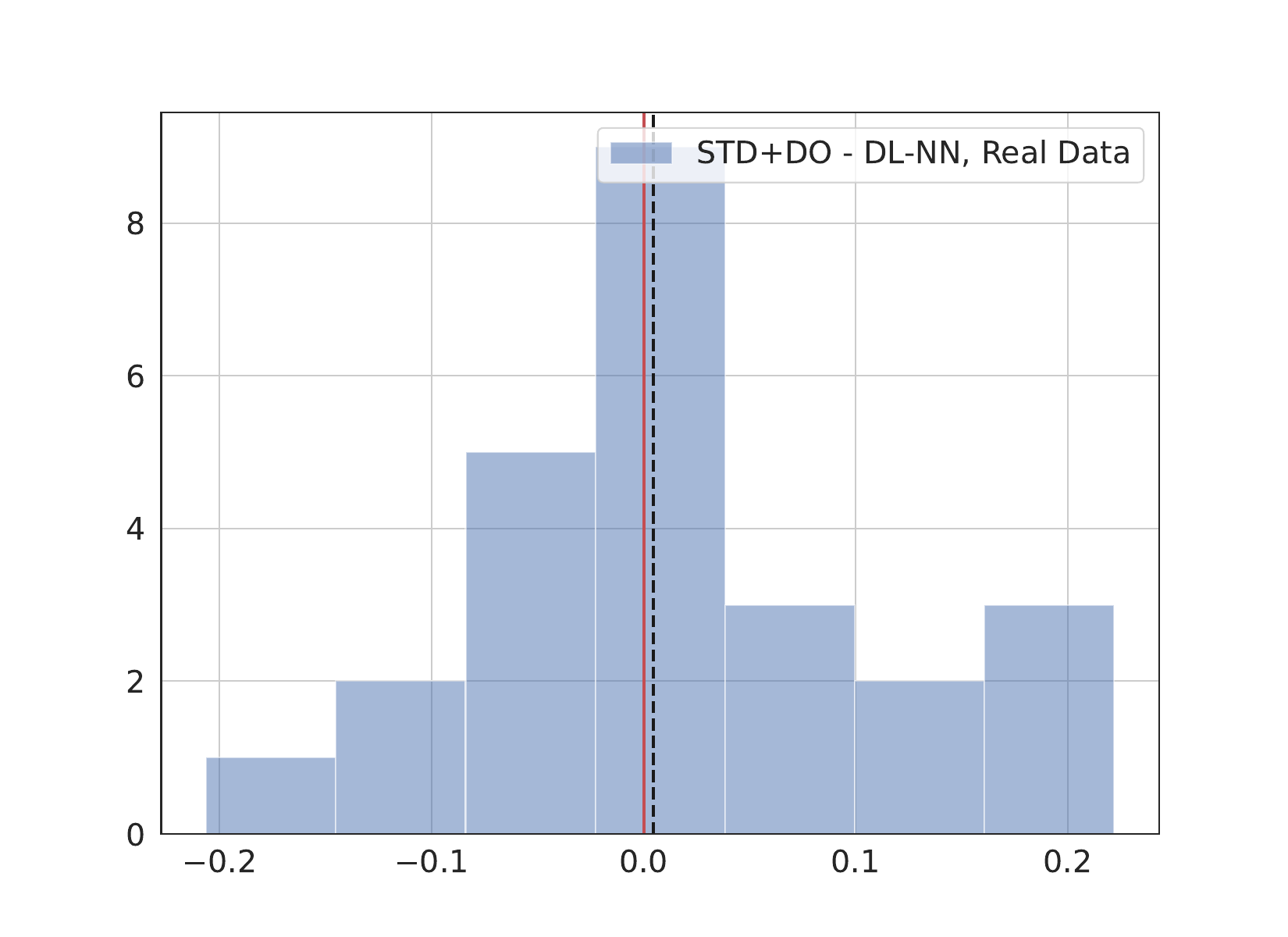}
  }
  \centerline{
  \includegraphics[trim={2cm 0cm 3cm 2cm},clip, scale=0.25]{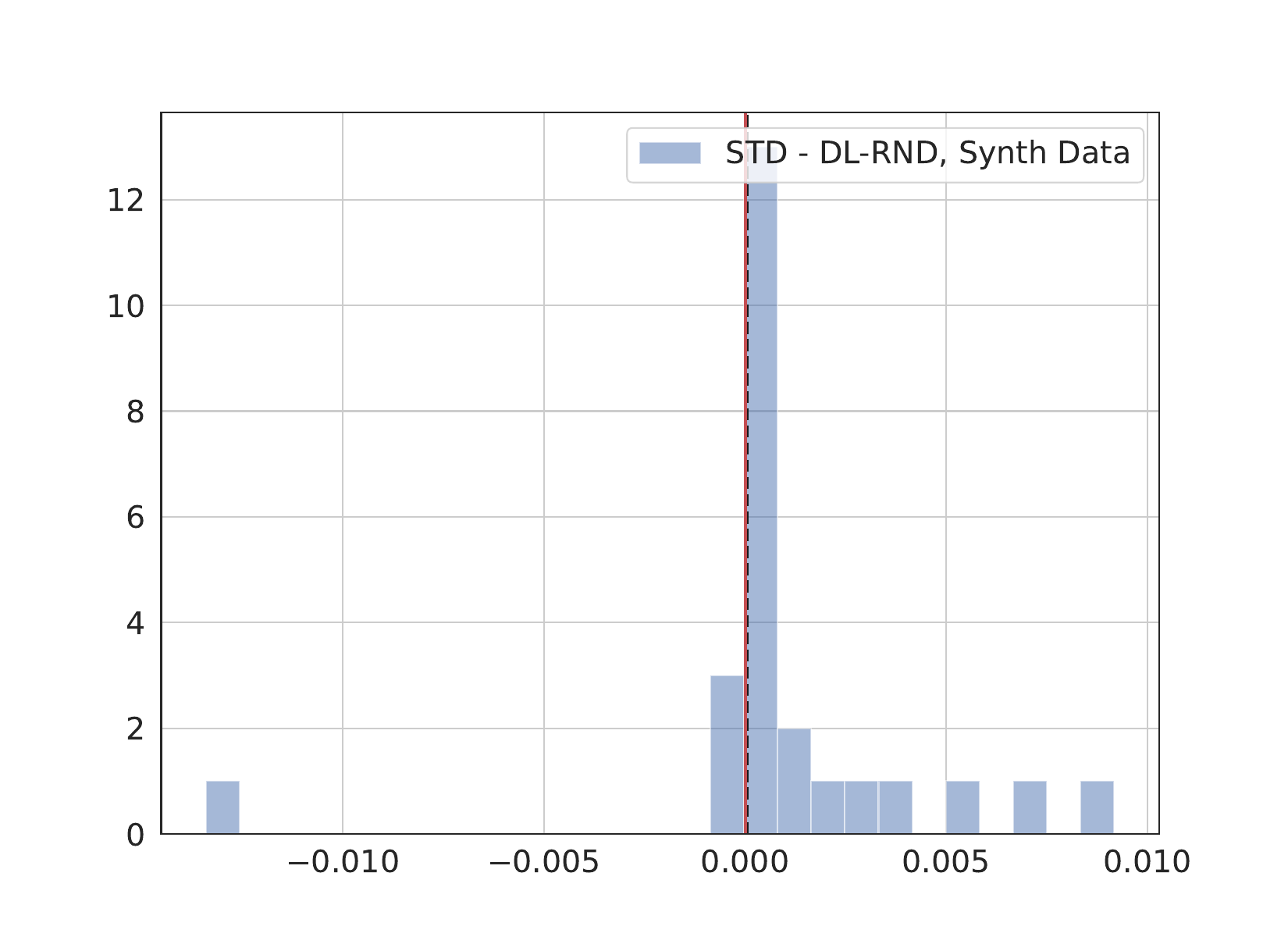}
  \includegraphics[trim={2cm 0cm 3cm 2cm},clip, scale=0.25]{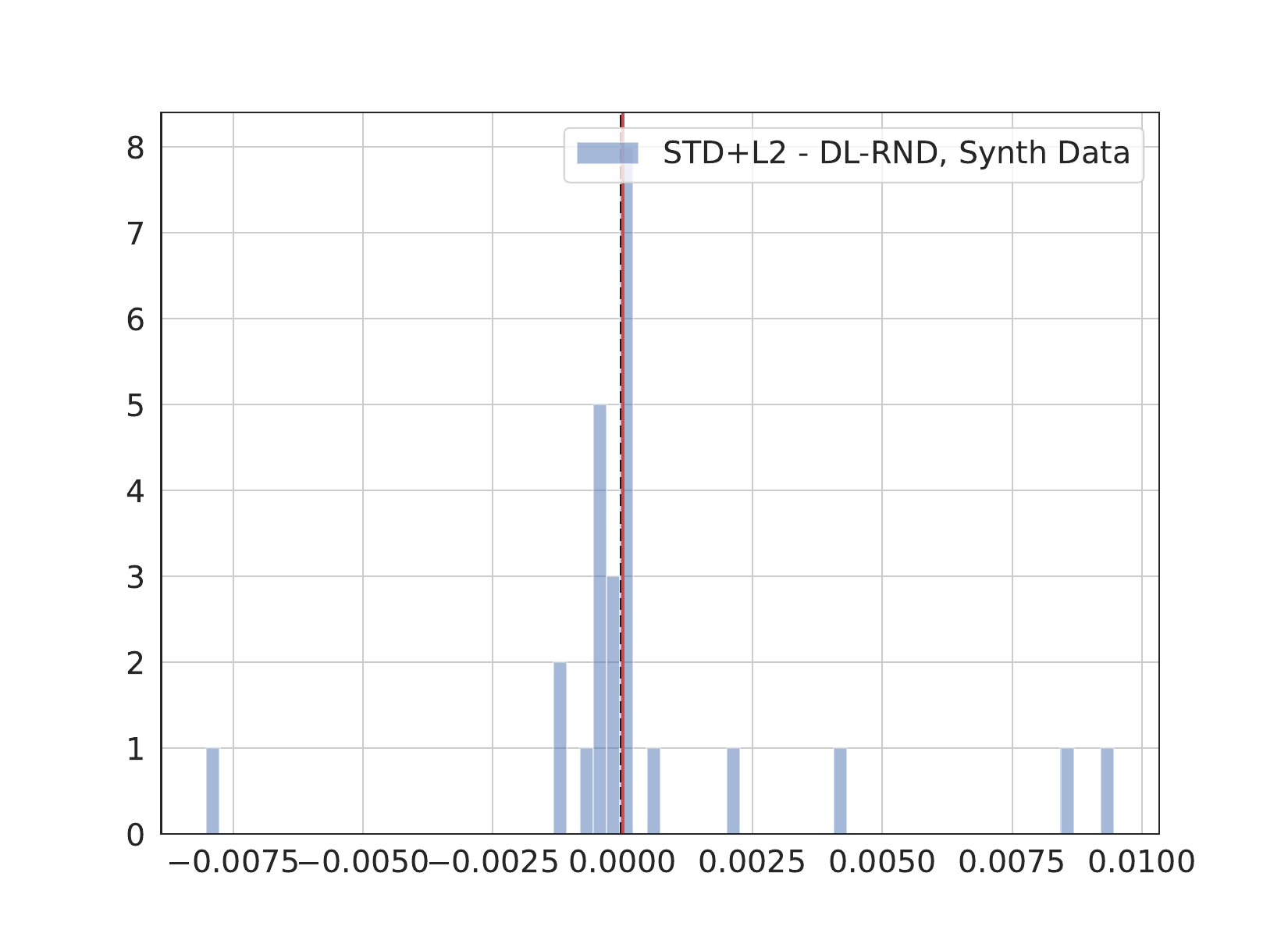}
  \includegraphics[trim={2cm 0cm 3cm 2cm},clip, scale=0.25]{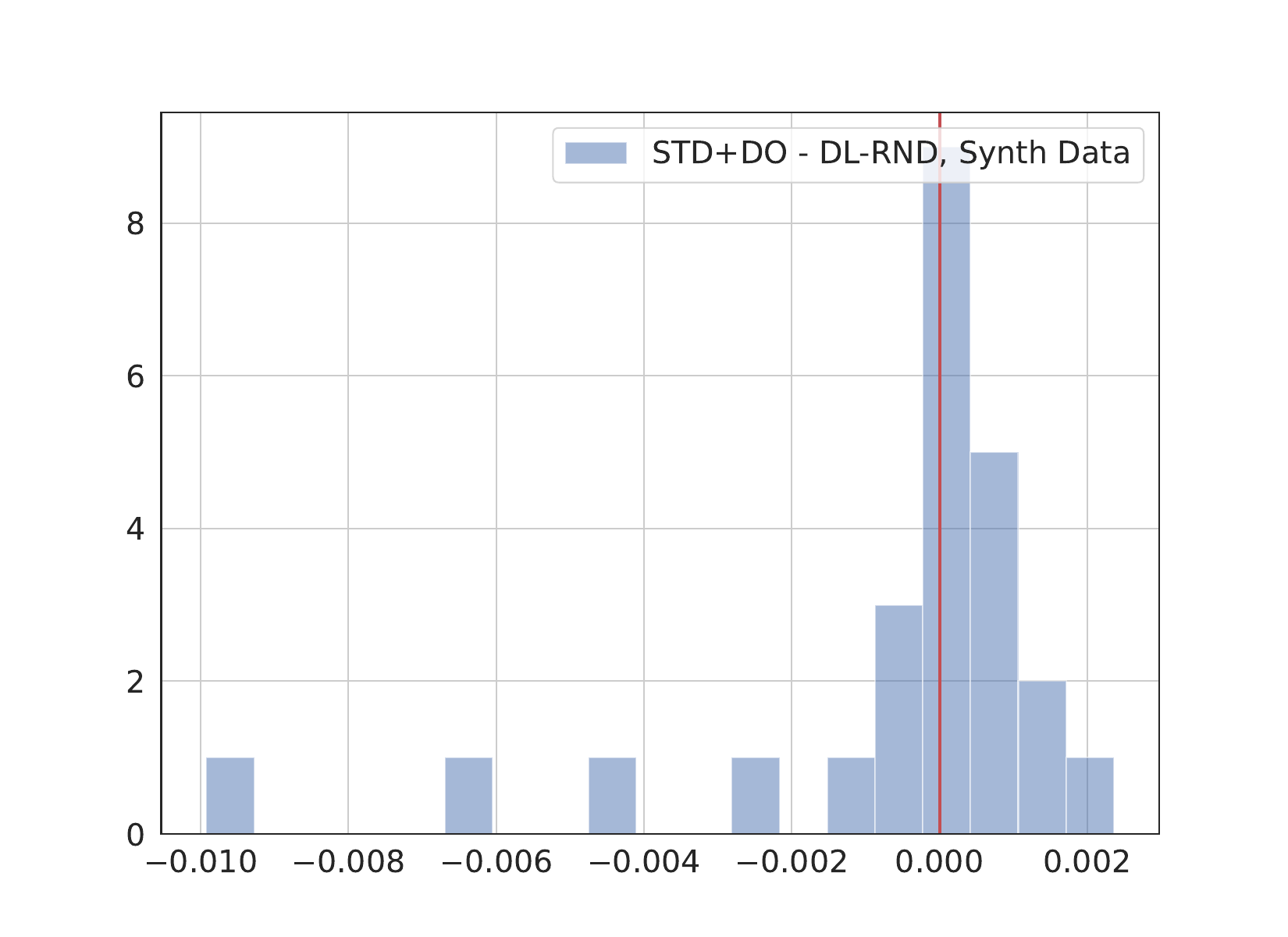}
 
  }
  \centerline{
  \includegraphics[trim={2cm 0cm 3cm 2cm},clip, scale=0.25]{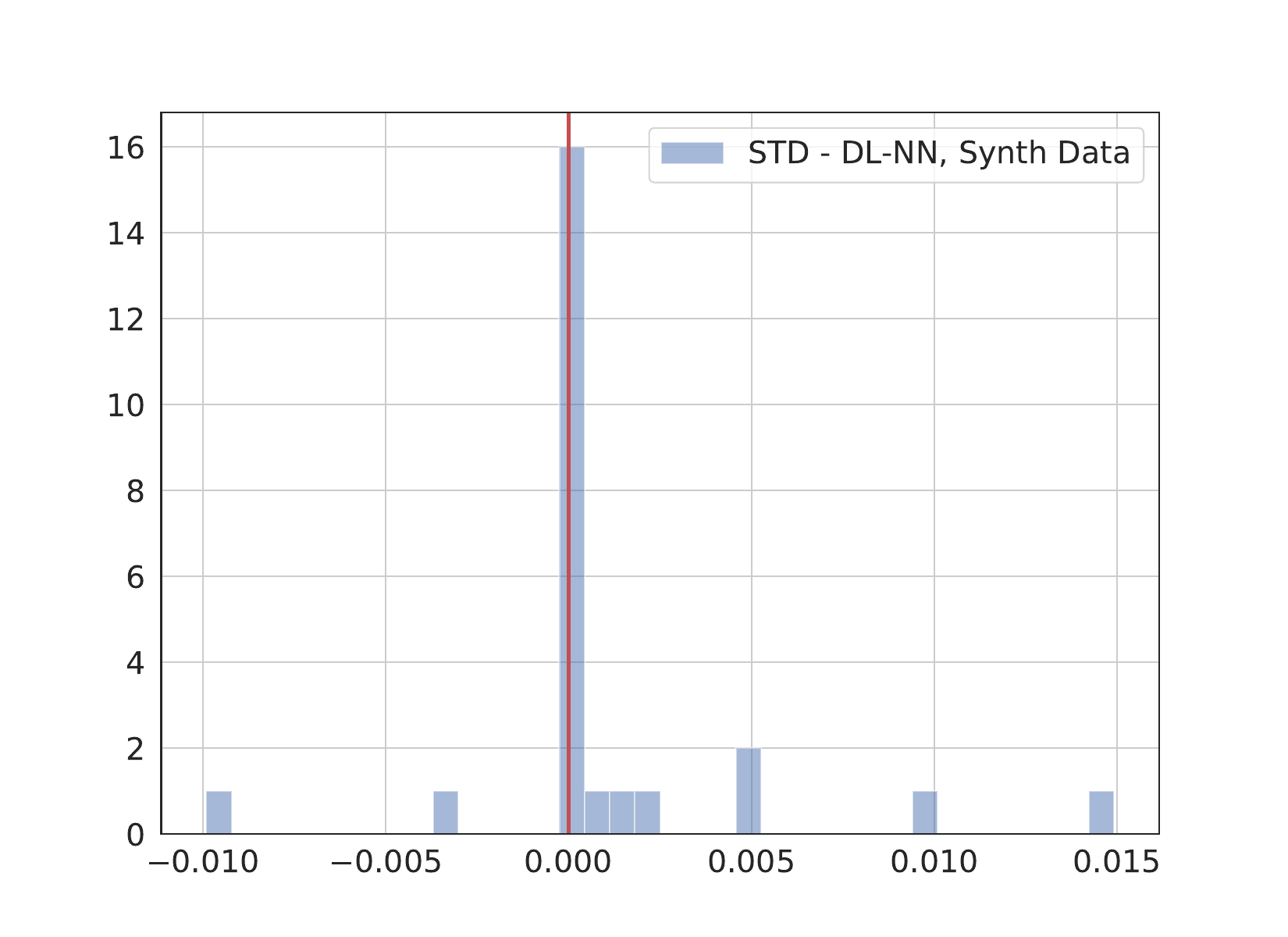}
  \includegraphics[trim={2cm 0cm 3cm 2cm},clip, scale=0.25]{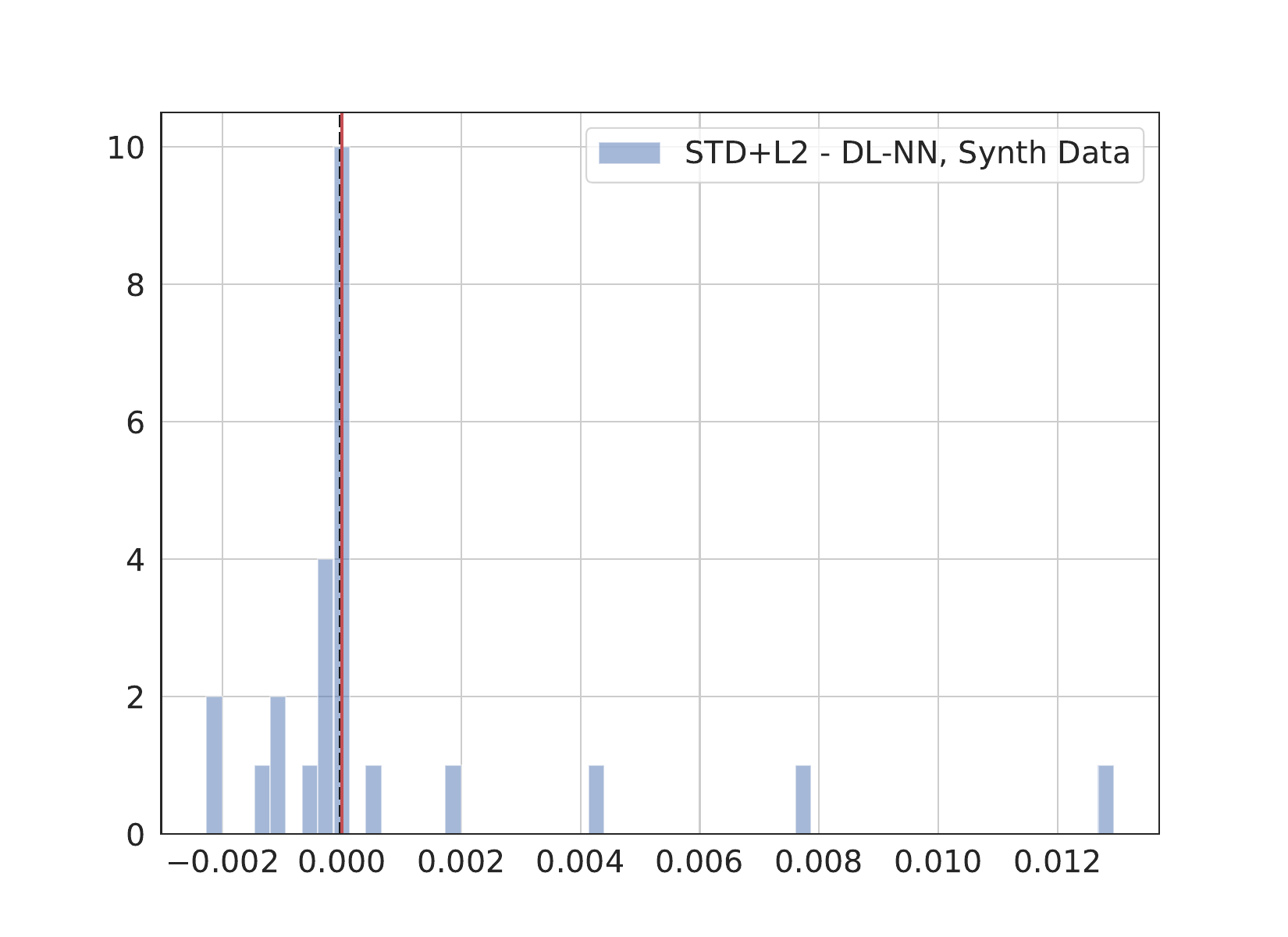}
  \includegraphics[trim={2cm 0cm 3cm 2cm},clip, scale=0.25]{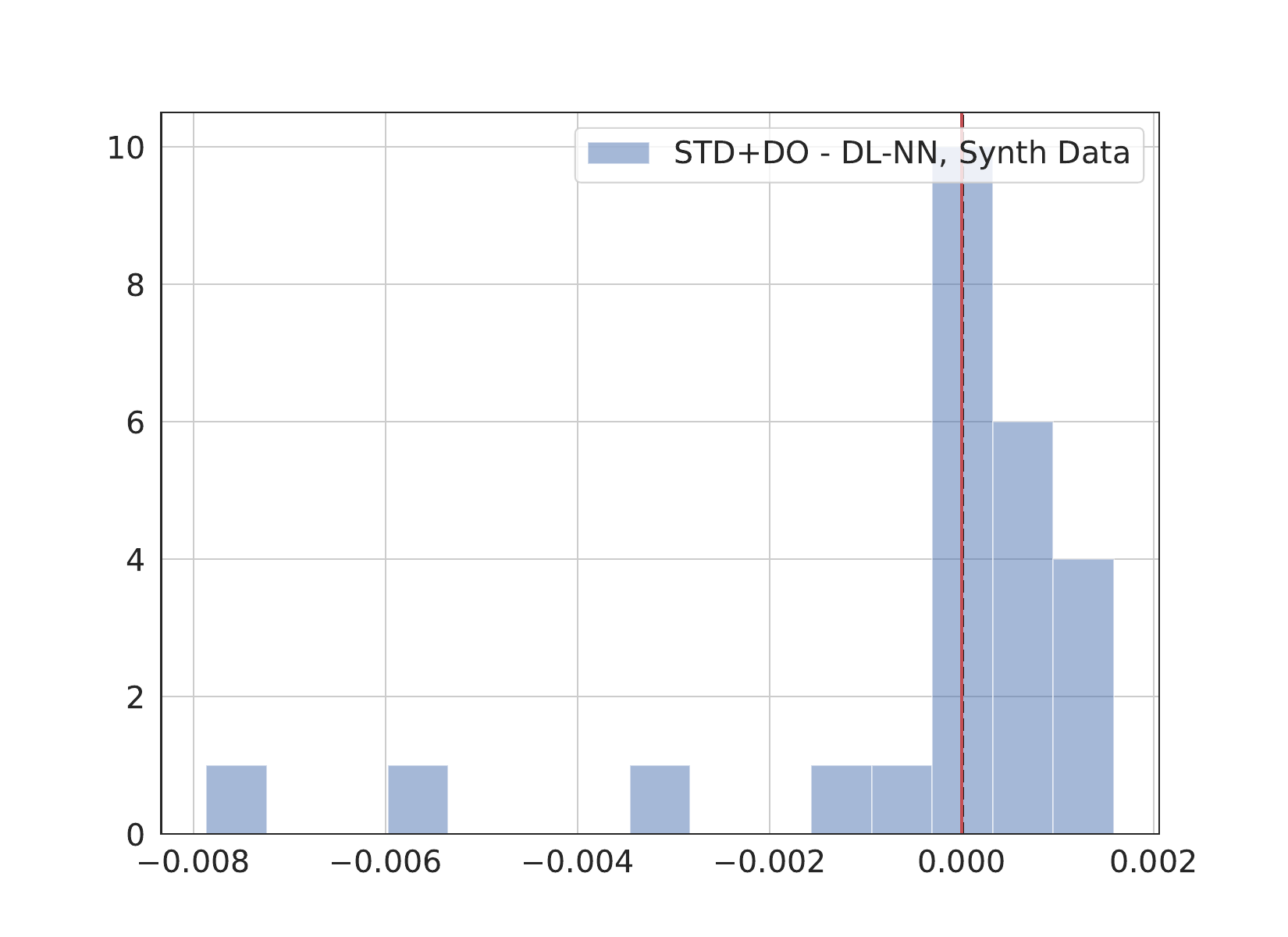}
  }
  \caption{The histograms of the pairwise difference of $MSE_{val}$ between the $STD$ groups ($STD$, $STD+L_2$, $STD+DO$) and the $DL$ groups ($DL_{RND}$,$DL_{NN}$) per each dataset and cross validation fold.
  The frequency reported is for each of the 5-folds belonging to the 5 real dataset (top 2 rows) and 5 synthetic dataset (bottom 2 row), for a total of 25 samples for each pairwise comparison.
  $Median_{\Delta}$ is the median of the differences in each comparison.
  We plot the $Median_{\Delta}$ as a dotted black line and the 0 line in red as a reference point. 
  For the synthetic data, the $Median_{\Delta}$ is so close to 0 that the two lines are not visually distinguishable. 
  }
  \label{fig:hist}

  \end{center}
\end{figure}

\appendix

%%%%%%%%%%%%%%%%%%%%%%%%%%%%%%%%%%%%%%%%%%%%%%%%%%%%%%%%%%%%%%%%%%%%%%%%%%%%%%%

\end{document}